\pdfoutput=1

\documentclass[11pt]{article}

\usepackage{acl}

\usepackage{times}
\usepackage{latexsym}

\usepackage[T1]{fontenc}

\usepackage[utf8]{inputenc}

\usepackage{microtype}

\usepackage{inconsolata}

\usepackage{xcolor}
\usepackage{graphicx}
\usepackage{amsmath}
\usepackage{bbm}
\usepackage{algorithm}
\usepackage{algorithmic}
\usepackage{hyperref}
\newcommand{\ours}{{RichRAG}}

\usepackage{graphicx}
\usepackage{hyperref}
\usepackage{array}

\usepackage{diagbox}
\usepackage{makecell}
\usepackage{arydshln} 
\usepackage{makecell} 
\usepackage{booktabs} 

\usepackage[most]{tcolorbox}
\usepackage{tikz}
\usepackage{tabularx}
\usepackage{subfigure,graphicx}
\usepackage{booktabs}
\usepackage{multirow}
\usepackage{makecell}
\usepackage{xcolor}
\usepackage{eqparbox} 
\usepackage{transparent}

\newcommand{\ie}{\textit{i.e.}}
\newcommand{\eg}{\textit{e.g.}}

\definecolor{mygreena}{RGB}{0,95,9}
\definecolor{myorange}{RGB}{170,82,0}
\definecolor{myblue}{RGB}{0,91,95}

\tcbset{
	mybox1/.style={
		colback=black!2!white, 
		colframe=black, 
		colbacktitle=gray, 
		coltitle=white, 
		boxsep=5pt, 
		left=5mm, 
		right=5mm, 
		top=5mm, 
		bottom=5mm, 
		arc=2mm, 
		auto outer arc, 
		boxrule=1pt, 
        title={GPT4 Prompt: Annotate question aspects and split long-form answers into corresponding sub-answers.}, 
	}
}

\title{RichRAG: Crafting Rich Responses for Multi-faceted Queries in Retrieval-Augmented Generation}

\author{Shuting Wang$^1$, Xin Yu$^2$, Mang Wang$^2$, Weipeng Chen$^2$, Yutao Zhu$^1$ \and \textbf{Zhicheng Dou$^{1*}$} \\
$^1$Gaoling School of Artificial Intelligence, Renmin University of China \\
$^2$Baichuan Intelligent Technology\\
\texttt{\{wangshuting, dou\}@ruc.edu.cn}
}

\begin{document}
\maketitle
\def\thefootnote{\arabic{footnote}}
\begin{abstract}
Retrieval-augmented generation (RAG) effectively addresses issues of static knowledge and hallucination in large language models. Existing studies mostly focus on question scenarios with clear user intents and concise answers. However, it is prevalent that users issue broad, open-ended queries with diverse sub-intents, for which they desire rich and long-form answers covering multiple relevant aspects. 
To tackle this important yet underexplored problem, we propose a novel RAG framework, namely \ours{}. It includes a sub-aspect explorer to identify potential sub-aspects of input questions, a multi-faceted retriever to build a candidate pool of diverse external documents related to these sub-aspects, and a generative list-wise ranker, which is a key module to provide the top-k most valuable documents for the final generator. These ranked documents sufficiently cover various query aspects and are aware of the generator's preferences, hence incentivizing it to produce rich and comprehensive responses for users. 
The training of our ranker involves a supervised fine-tuning stage to ensure the basic coverage of documents, and a reinforcement learning stage to align downstream LLM's preferences to the ranking of documents. Experimental results on two publicly available datasets prove that our framework effectively and efficiently provides comprehensive and satisfying responses to users.
\end{abstract}

\section{Introduction}
\label{sec:intro}
\begin{figure}
    \centering
    \includegraphics[width=0.95\linewidth]{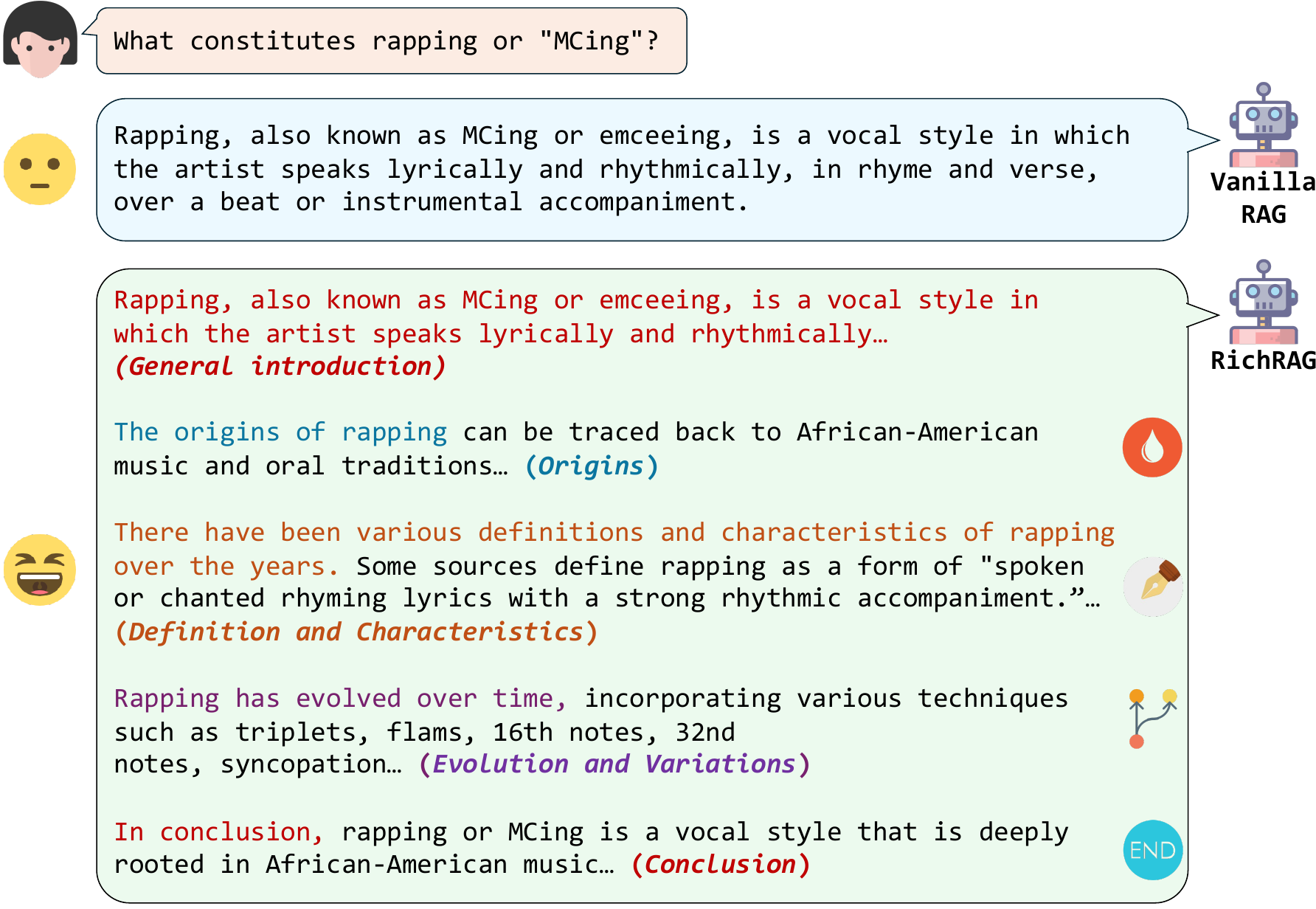}
    \caption{An example of a scenario where a multi-faceted query requires a comprehensive answer.}
    \label{fig:intro}
    \vspace{-10pt}
\end{figure}

Large language models (LLMs) have revolutionized how information is accessed online, shifting from returning ranked lists of relevant documents to directly generating answers to user queries. However, they still suffer from hallucinations and information staleness issues, impacting the authenticity and reliability of generated answers. Retrieval-augmented generation (RAG) has emerged as a promising solution, empowering LLMs to leverage reliable information from retrieved documents, thereby returning more reliable responses. 

Though some advanced techniques~\cite{flare,self-rag,zhengbao_filter,xiong_filter} have been proposed, existing studies primarily focus on addressing specific problems that require concise and definitive answers. However, user intents are complex and multi-faceted, necessitating rich and comprehensive answers. As Figure~\ref{fig:intro} shows, when a user inquires about rapping-related information, a rich response about various aspects of rapping, such as origins, characteristics, and evolution could lead to a more satisfactory user experience than a superficial description.

Our research is focused on developing effective RAG approaches to handle these more complex user needs.
We propose a RAG framework, \ours{}, which is designed to offer diverse external knowledge that comprehensively covers various sub-aspects of multi-faceted queries, thereby enhancing the downstream generator (an LLM) to yield rich responses. \ours{} first employs a sub-aspect explorer to explicitly predict sub-aspects of queries. Then, it adopts a multi-facet retriever to build a broad pool of candidate documents covering those identified sub-aspects. However, such redundant candidates inevitably contain much irrelevant noise and are hard to handle completely by LLMs due to limited input length. As a result, sorting out the top-k best documents from the candidate pool is critical to the success of the \ours{} framework.

In further, we claim that a promising top-k ranking should have the following desirable features: 
(1) \textit{Comprehensiveness}. Incentivizing LLM to generate rich and reliable responses requires external documents to comprehensively cover various query aspects. 
Therefore, instead of predicting each document's relevance independently, the ranking model should consider the relationship among documents to enhance the global coverage of the ranking list for query aspects.
(2) \textit{Alignment with the LLMs' preferences}. In RAG systems, the users of IR models are LLMs instead of humans. Thus, the reference order should be LLM-friendly, hence enhancing the generator to produce satisfying responses.

To achieve this, we devise a generative list-wise ranker based on encoder-decoder structures. It takes as input the user query, its identified sub-aspects, and all candidates, then directly generates top-k document IDs as final ranking lists. This structure offers two key advantages:
(1) \textit{Global Document Modeling}. 
The seq-to-seq model structure equips the ranker to 
effectively model global interactions among candidates, queries, and sub-aspects, thereby capturing the overall utility of generated ranking lists in covering the query's multi-aspects. 
(2) \textit{Efficiency}. Following the FiD structure~\cite{fid}, we parallelize the encoding of each candidate and further introduce pooling and reuse operations to the decoder module. These strategies significantly reduce the spatiotemporal load of the ranker.

The optimization of our ranker consists of two stages:
The first is supervised fine-tuning (SFT). 
To enhance the coverage of generated ranking lists on query aspects, we devise a \textit{coverage utility function} based on which to build silver generation targets (ranking lists of document IDs) for training samples greedily. These silver targets allow us to SFT our ranker and ensure its basic ability.
To further improve the ranking quality and align ranking lists with LLMs' preferences, a reinforcement learning stage is introduced. We consider both the accuracy and comprehensiveness of generated responses to create reward values, and adopt the DPO~\cite{DPO} algorithm to optimize our ranker. 
In addition, we devise a \textit{unilateral significance sampling strategy} (US$^3$) to build valuable training samples for stable optimization.  
Experiments on two public datasets prove that \ours{} can effectively and efficiently generate more comprehensive answers for multi-faceted queries than existing methods. 

Our contributions are three-fold:

(1) We propose a RAG framework \ours{} to explicitly model the query's various sub-aspects, thereby providing comprehensive long-form responses to satisfy the user's rich intents.

(2) We develop an efficient generative list-wise ranker that models the global gain of ranking lists considering rich user intents, delivering promising ranking lists for downstream LLMs.  

(3) We devise the US$^3$ approach to create reliable and valuable training pairs for the DPO algorithm, improving the quality and stability of optimization.

\section{Related Works}
\subsection{Retrieval-augmented Generation}
To ensure the effectiveness of RAG systems, previous studies mainly optimized retrievers and generators simultaneously~\cite{atlas,retro,retroplus,realm,hindsight,rag,radit,self-rag}. 
Recent researchers also explored fixing LLMs and optimizing retrievers as plug-in modules~\cite{replug,ARR} or introducing post-retrieval components, \eg, compressors and rankers~\cite{comp1,zhengbao_filter,xiong_filter,bgm,ragsurvey}, to reduce the training cost. 
Some studies~\cite{chan2024rqrag,wang2024selfdc,khot2023decomposed,yao2023react,searchchain} propose to decompose multi-hop questions into sub-tasks and solve them step-by-step. These works typically focus on breaking down \textit{complex questions with clearly stated} user intents into simpler questions. Our research, however, addresses a different scenario where user questions are broad and encompass \textit{various potential sub-aspects not explicitly stated}. Answering such questions usually requires integrating diverse relevant information to these underlying sub-aspects to fully respond to the user's potential information needs.
Recent studies~\cite{ragfusion} also highlight the importance of exploring users' sub-intents, but its simplistic pipeline fails to model global document-intent interactions, leading to sub-optimal results.

\subsection{Generative Ranking with LLM}
Recently, the rise of LLMs allows researchers to establish various generative ranking models~\cite{rankt5-orig,rankt5,rankgpt,lit5,pair1,pair2}, including point-wise~\cite{rankt5-orig,rankt5}, pair-wise~\cite{pair1,pair2}, and list-wise~\cite{lit5,rankgpt} models. 
To handle extensive document load, some generative list-wise methods~\cite{rankgpt,lit5} adopt a sliding-window approach to iteratively generate final ranking lists.
Some studies~\cite{bgm,xu2024listaware} also explore list-wise rankers in RAG systems but still focus on scenarios with specific intents and answers, neglecting the depth and breadth of user questions. 

\section{Method}\label{sec:method} 

\begin{figure*}
    \centering
    \includegraphics[width=0.85\linewidth]{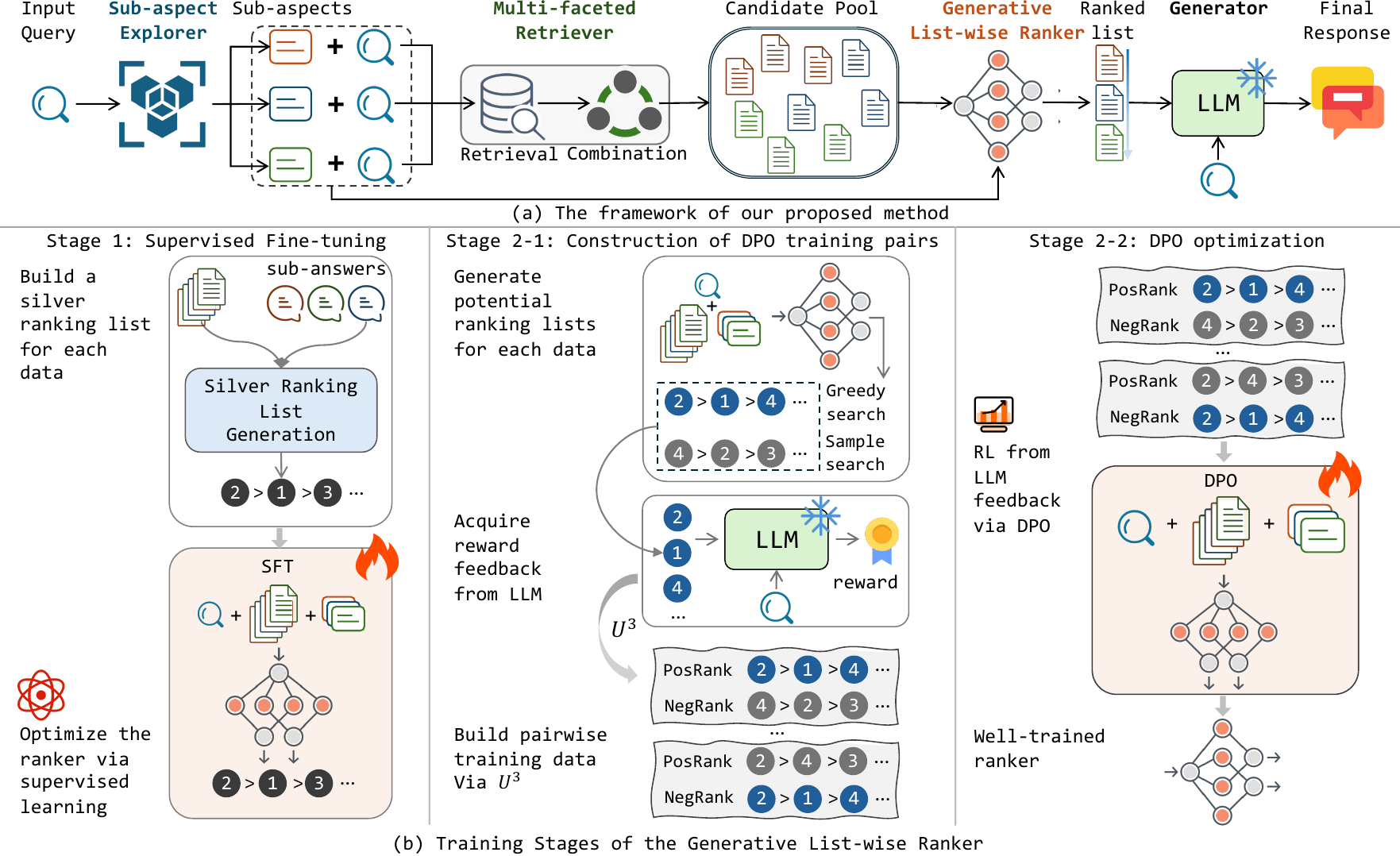}
    \caption{The overall framework of \ours{}. We describe the training stages of our ranker at the bottom.}
    \label{fig:framework}
\end{figure*}

In this section, we demonstrate our \ours{} framework, which explicitly considers the sub-aspects of multi-faceted questions to provide diverse and LLM-friendly external reference lists, thereby enhancing the richness and satisfaction of generated responses. Figure~\ref{fig:framework} (a) displays the overall framework. 
We first define the problem, and then delve into the introduction of each component, including the sub-aspect explorer, the multi-faceted retriever, and the generative list-wise ranker.
\subsection{Problem Definition}\label{sec:problem-definition}
The basic RAG setting usually contains a knowledge corpus $\mathcal{C}$, a fixed retriever $\mathcal{R}$, and a fixed LLM serving as the generator, $\mathcal{G}$. For a multi-faceted query, $q$, its various subordinate aspects are denoted as $\mathcal{S}=\{s_1,\ldots, s_n\}$. 
These sub-aspects have corresponding sub-answers, which are denoted as $\mathcal{A}=\{a_1,\ldots,a_n\}$.\footnote{The collection of sub-aspects and sub-answers is introduced in Appendix~\ref{app:dataprocess}.} The combination of these sub-answers forms the ground truth answer, $a$, which is long-form and responds to all sub-aspects. Existing RAG models primarily focus on retrieving relevant documents from the corpus and incorporating them into the LLM's input to generate responses closely aligned with ground truth answers. In this study, we aim to make responses, $r$, generated by \ours{} not only match the ground truth answers but also sufficiently cover individual sub-answers comprehensively, to ensure the responses' richness and completeness.

\subsection{Sub-aspect Explorer}\label{sec:se}
Examining various sub-aspects under a user's query could provide explicit insights into the user's underlying intents, thereby enabling more satisfactory results for users~\cite{xquad,pm2,DVGAN,expliPS}. 
We leverage LLMs to build our sub-aspect explorer, $\mathcal{E}$, due to their extensive world knowledge and excellent capabilities in language understanding and generation. 
This module takes a prompt $p_{se}$, which instructs the LLM to predict the sub-aspects of the input query, and a user's query, $q$, as input and generates a series of sub-aspects under the query: 
\begin{equation}\small\vspace{-2pt}
     {\hat{\mathcal{S}}}=\{\hat{s}_1,\ldots,\hat{s}_m\} = \mathcal{E}(q,p_{se}).
\end{equation}\vspace{-2pt}
To align the sub-aspect explorer with the output format and the distribution of downstream data, we fine-tune it using training queries and their labeled sub-aspects. The target output is a concatenation of labeled sub-aspects surrounded by square brackets: $o = [s_1]\ldots[s_n].$ Subsequently, we optimize the sub-aspect explorer by the next token prediction (NTP) loss function:
\begin{equation}\small
\begin{aligned}
    \mathcal{L}_{se}=-\sum\nolimits_{i=1}^{|o|}\log P(o_i|o_{1:i-1}, q, p_{se}).
\end{aligned}\vspace{-1pt}
\end{equation}

\subsection{Multi-faceted Retriever}\label{sec:retriever}
Given the query's sub-aspects that represent the user's various potential sub-intents, we then use a multi-faceted retriever to collect documents that are relevant to various sub-aspects to build a diverse candidate pool. This operation could filter out apparent irrelevant documents and shrink the search space of the subsequent ranker. The multi-faceted retriever consists of the following two processes.

The first is a retrieval process, where we separately retrieve top-N documents $\mathcal{D}_i$ from the corpus for each sub-aspect $\hat{s_i}$. 
To avoid the topic drift, we concatenate each sub-aspect with the original query to form a new query and retrieve as follows, 
\begin{equation}\small
    \mathcal{D}_i = \mathcal{R}(q\circ \hat{s_i}, \mathcal{C}), 
\end{equation}\vspace{-2pt}
where $\circ$ denotes concatenation and each document in $\mathcal{D}_i$ is associated with the sub-aspect, $\hat{s_i}$. 

Next, a combination process is introduced to merge all these retrieved documents to create the candidate pool, $\mathcal{P}$. Since some documents may be retrieved multiple times by different sub-aspects, to reduce the space-time burden of the ranker, we treat the repeated documents as a single one, hence the associated sub-aspects form a set, $s(d)$: 
\begin{equation}\small
    \begin{aligned}
         \mathcal{P} &= \mathcal{M}(\mathcal{D}_{1:m})=\{d_1,\ldots,d_M\},\\
        s(d_i) &= \{\hat{s}^i_1, \ldots,\hat{s}^i_{n_i}\}.
    \end{aligned}\vspace{-2pt}
\end{equation}
$\mathcal{M}()$ denotes the combination and $M$ is the maximum capacity of the pool. This candidate pool collects potentially valuable documents that sufficiently cover various sub-aspects of the query.

\subsection{Generative List-wise Ranker}\label{sec:franker}
Though we have collected plenty of candidates related to various sub-aspects, directly providing these massive documents to the generator is challenging due to the extensive processing burden and potential noisy information. 
Consequently, we devise a ranking model that targets to sort out the top-$k$ most valuable documents from the candidate pool. These ranked documents should collectively cover the query's various sub-aspects and adhere to the preferences of the generator, hence enhancing the response performance. 
To equip our ranker with the ability to globally model relationships among candidates, we build it upon a generative model, 
T5~\cite{t5}, which views all candidates, sub-aspects, and the query as input and directly generates a top-k ranking list of document ID (docid) tokens. For each candidate, $d_i$, we concatenate it with the original query, $q$, its associated sub-aspects, $s(d_i)=\{\hat{s}^i_1, \ldots,\hat{s}^i_{n_i}\}$, and some special tokens to formulate an input sequence:
\begin{equation}\small
    I_i = [\mathrm{D}_i]\circ q\circ[\mathrm{Q}]\circ\hat{s}^i_1\circ[\mathrm{E}]\ldots\;\hat{s}^i_{n_i}\circ[\mathrm{S}]\circ d_i,
\end{equation}\vspace{-2pt}
where $[\mathrm{D}_i]$ is the docid token indicating the $i$-th candidate, $[\mathrm{Q}]$ denotes the end of the query, $[\mathrm{E}]$ separates each sub-aspect, and $[\mathrm{S}]$ denotes the end of associated sub-aspects. 
Inspired by FiD structure~\cite{fid}, the encoder module, $\mathrm{Enc}()$, encodes candidate sequences in parallel to ensure high efficiency. Furthermore, since the ranker's generation space is limited to docid tokens instead of the whole vocabulary, 
we use the pooling operation $\mathrm{Pool}()$ to extract the encoded output of docid tokens $\mathrm{\mathbf{e}}_i$ as relevance representations of candidates. They are then connected and entered into the decoder $\mathrm{Dec}()$ to generate the ranking list: 
\begin{equation}\small
\begin{aligned}
    [\mathrm{D}_{r(1)}], \ldots, [\mathrm{D}_{r(k)}] &= \mathrm{Dec}([\mathrm{\mathbf{e}}_1;\ldots;\mathrm{\mathbf{e}}_M]), \\
    \mathrm{\mathbf{e}}_i &= \mathrm{Pool}(\mathrm{Enc}(I_i)),
\end{aligned}
\end{equation}
where $r(i)$ project the rank position $i$ into the index of the document ranked at the $i$-th position. This pooling operation could significantly reduce the time-space burden of the decoder. 


Additionally, we implement a reuse strategy on the language model (LM) head layer to reduce unnecessary load and enhance the modeling accuracy. It sets the LM head layer's projection matrix $\mathrm{\mathbf{F}}$, which maps generated hidden states to probabilistic token spaces, to be $d \times M$, ($M$ is the maximum number of candidates). Furthermore, our preliminary experiments imply that randomly initializing the value of $\mathrm{\mathbf{F}}$ is hard to optimize due to limited training samples. Therefore, we define its value using relevance representations of candidates to simplify optimization difficulty, hence improving the ranking performance. Thus, the probability of the $t$-th token is computed as below, 
\begin{equation}\small
\begin{aligned}
    p^t(d)=\mathrm{Softmax}\left(\mathbbm{M}(\mathrm{\mathbf{h}}^t\mathrm{\mathbf{F}}/{\tau})\right),\mathrm{\mathbf{F}} = [\mathrm{\mathbf{e}}_1;\ldots;\mathrm{\mathbf{e}}_M],
\end{aligned}
\end{equation}
where $\mathrm{\mathbf{h}}^t$ is generated hidden states of the $t$-th token and $\tau$ is temperature to control the sharpness of distribution.
$\mathbbm{M}()$ denotes a masking mechanism to set the probabilities of previously generated documents to zero, avoiding repetition. To ensure the ranker's performance, we employ a two-stage optimization. We demonstrate it in the following sections and visualize it in Figure~\ref{fig:framework} (b).

\subsubsection{Supervised Fine-tuning}\label{sec:sft}
To address the problem of the vast searching space of possible permutation, a major challenge for list-wise ranking algorithms, we adopt a greedy algorithm to build silver target ranking lists for each training instance, supporting supervised fine-tuning of the ranker. Specifically, we devise a coverage utility function, $\Phi^t(d)$, to measure the incremental gain in aspect coverage for each remaining document, $d$, conditioned on previously selected ones, $L^*_{t-1}$. The greedy selection is presented as follows,
\begin{equation}\small
    \begin{aligned}
        L^*_{t} &= L^*_{t-1} \cup d^*_t,\;\;d^*_t = \mathop{\arg\max}\nolimits_{d\in\mathcal{P}/L^*_{t-1}} \Phi^t(d),\\
        \Phi^t(d) &= \sum\nolimits_{i=1}^{n} w^t_i\phi(d, a_i).
    \end{aligned}
\end{equation}
$\Phi^t(d)$ considers the current importance of each sub-aspect $w^t_i$ and the candidate's coverage for each sub-aspect $\phi(d, a_i)$. 
The coverage function $\phi()$ is implemented by the rouge-score between $d$ and the $i$-th sub-answer $a_i$. The current importance of sub-aspects is measured by calculating their coverage by previous $t-1$ selected documents using the following function with sum normalization $\mathrm{Norm}()$:
\begin{equation}\small
    w^t_i = 1-\mathrm{Norm}_i({\max\nolimits_{\tilde{d}\in L^*_{t-1}}\phi(\tilde{d},a_i)}),\label{eq:sub-weight}
\end{equation}
The silver target list, $L^*_k$, allows us to supervise fine-tune the ranker via the NTP task:
\begin{equation}\small
    \mathcal{L}_{sft} = -\sum\nolimits_{t=1}^{k}\log p(d^*_t | q, \mathcal{P}, \hat{\mathcal{S}},L^*_{t-1}). 
\end{equation}
$p(d^*_t | q, \mathcal{P}, \hat{\mathcal{S}},d^*_{1:t-1})$ is the generation probability of the $t$-th target docid conditioned on the current question, the candidate pool, the sub-aspects, and the previously target documents. It is calculated by our ranking module.

\subsubsection{Reinforcement Learning}\label{sec:dpo}
After supervised fine-tuning, aligning ranking lists with LLM-preferred order is critical to enhance the final response quality. Therefore, we use an RL strategy to explore better ranking possibilities.

$\bullet$ \textit{Reward Function.}
We treat the quality of final responses as the reward of provided ranking lists to model LLM's preferences. Since we expect the generated responses to cover all sub-answers from all sub-aspects, besides using the rouge score $\phi()$ to calculate the matching between responses $r$ and golden answers $a$, we further introduce a com-rouge score, $\phi^c()$ to measure the coverage of responses on sub-answers, $\mathcal{A}$. The reward function $\mathcal{r}(L)$ of a ranking list $L$ is produced as:
\begin{equation}\small
\begin{aligned}
\label{eq:reward}
    \mathcal{r}(L) &= \phi(r, a)+\phi^c(r, \mathcal{A}), \\
    \phi^c(r, \mathcal{A}) &= \sum\nolimits_{i=1}^n\delta_i\phi(r, a_i),\;\;\; r = \mathcal{G}(q,L).
\end{aligned}
\end{equation}
$r$ is the response from the generator, $\mathcal{G}()$, given the query $q$ and ranked top-$k$ documents $L$. 
$\delta_i$ denotes the normalized length of the sub-answer to value the sub-answer's weight.


Then, we adopt the Direct Preference Optimization (DPO)~\cite{DPO} algorithm to ensure the optimization stability. Its training samples consist of a series of prediction pairs pre-generated by the policy model, namely the ranker in our study. Each pair contains a winner and a loser prediction, namely generated ranking lists, which are assessed by their rewards. Thus, DPO pairwise optimizes the policy model to discriminate the better one among a prediction pair. 

$\bullet$ \textit{Data Construction.} 
To build valuable training pairs, we introduce the \textit{unilateral sample significance strategy} (US$^3$). First, this approach generates a greedy search ranking list and multiple sampled ranking lists for each training data, obtaining their rewards via Equation~(\ref{eq:reward}). Then, US$^3$ follows two rules when forming DPO training pairs: (1) \textit{Unilaterality}: One prediction is from greedy search (used in inference) to provide a baseline for discerning better optimization directions, and the other from sampling search. (2) \textit{Significance}: The reward gap between the predictions must exceed a threshold $\mu$ to ensure the pair's value, thereby reducing errors from pairs with similar performance that may not reflect ranking quality, but noise.


$\bullet$ \textit{Optimization.} Given built training pairs, we optimize the ranker using the following DPO objective function:
\begin{equation}\small
\begin{aligned}
    \mathcal{L}_{DPO} = -\!\!\!\!\!\!\!\!\!\mathop{\mathbbm{E}}_{(x,y_w,y_l)\sim D}\!\!\!\!\!\![\log \sigma (\beta \log\frac{\pi_\theta(y_w|x){\pi_{f}(y_l|x)}}{\pi_{f}(y_w|x)\pi_\theta(y_l|x)}
     ].
\end{aligned}
\end{equation}
$D$ is the training set built by US$^3$, where the input of each data is $x=\{q, \hat{\mathcal{S}}, \mathcal{P}\}$ and the output is a pair of ranking lists, with $y_w$ and $y_l$ as winner list and loser list respectively. $\pi_\theta$ denotes the policy model that needs to be optimized and $\pi_{f}$ represents the original policy model with fixed parameters, and its role is to avoid optimization trajectory excessively deviating from the basic model.

\begin{table*}[!t]
\small
    \centering
    
    \setlength{\tabcolsep}{1.6mm}{
    \begin{tabular}{llcccccccccccc}
    \toprule
    \multirow{2}*{Settings} & \multirow{2}*{Models} & \multicolumn{6}{c}{WikiPassageQA} &  \multicolumn{6}{c}{WikiAsp} \\
    \cmidrule(r){3-8}\cmidrule(l){9-14}
    & & {F1} & {R2} & {RL} & {BS} & {CR2} & {CRL} & {F1} & {R2} & {RL} & {BS} & {CR2} & {CRL} \\
    \midrule
    \multirow{13}*{\makecell{Predicted \\Sub-aspects}} 
    & Close-book & .2132   & .0324 & .0989 & .8417  & .0397 & .1714 & .2037 & .0114 & .0581 & .7967 & .0128 & .0834  \\
    \cmidrule{2-14}
    & No Ranker & .3372 & .1191 & .2065 & .8491  & .1310 & .2969 & 0.3029 & .0479 & .0923 & .8073  & .0502 & .1257  \\
    & RankT5    & \underline{.3502} & .1313 & .2123 & .8536  & .1441 & .3027 & 0.3079 & .0481 & .0932 & .8063  & .0504 & .1263  \\
    & LDIST     & .3487 & \underline{.1325} & \underline{.2140} & .8523  & \underline{.1447} & \underline{.3044} & .3377 & .0479 & .0923 & .8114 & \underline{.0624} & .1376  \\
    & LiT5      & .3473 & .1291 & .2118 & .8514 & .1413 & .3033 & .3341 & \underline{.0594} & \underline{.1037} & .8125 & .0620 & \underline{.1399}  \\
    \cmidrule{2-14}
    & RAG-Fusion    & .3316 & .1146 & .2029 & .8467 & .1261 & .2913 & .3219 & .0547 & .0976 & .8116 & .0572 & .1324  \\
    & \quad +RankT5 & .3448 & .1278 & .2122 & .8510 & .1402 & .3015 & .3123 & .0507 & .0939 & .8091 & .0532 & .1277  \\
    & \quad +LDIST  & .3400 & .1253 & .2114 & .8494 & .1394 & .3022 & .3327 & .0587 & .1021 & .8122 & .0612 & .1380  \\
    & \quad +LiT5   & .3386 & .1226 & .2071 & .8491 & .1347 & .2987 & .3344 & .0589 & .1017 & \underline{.8137} & .0616 & .1373  \\
    \cmidrule{2-14}
    & BGM    & .3465 & .1191 & .2065 & \underline{.8547}  & .1537 & .2983 & .2969 & .0472 & .0920 & .8061 & .0497 & .1242 \\
    & \ours{} & \textbf{.3638} & \textbf{.1538} & \textbf{.2316} & \textbf{.8549} & \textbf{.1664} & \textbf{.3225} & \textbf{.3610} & \textbf{.0678} & \textbf{.1094} & \textbf{.8194} & \textbf{.0706} & \textbf{.1458}  \\
    \midrule
    \multirow{12}*{\makecell{Golden \\Sub-aspects}} 
    & No Ranker    & .3763 & .1667 & .2650 & .8544 & .1809 & .3600 & .3325 & .0655 & .1277 & .8116 & .0688 & .1671  \\
    & RankT5       & .3854 & .1760 & .2734 & .8569 & .1887 & .3673 & .3329 & .0632 & .1244 & .8107 & .0665 & .1637  \\
    & LDIST        & .3867 & .1766 & .2728 & .8558 & .1906 & .3680 & .3609 & .0792 & .1354 & .8163 & .0826 & .1767  \\
    & LiT5         & \underline{.3926} & .1844 & .2808 & .8558 & .1991 & .3724 & .3564 & \underline{.0801} & \underline{.1385} & .8165 & \underline{.0834} & \underline{.1795}  \\
    \cmidrule{2-14}
    & RAG-Fusion    & .3808 & .1764 & .2740 & .8554 & .1901 & .3687 & .3658 & .0790 & .1376 & .8187 & .0825 & .1762  \\
    & \quad +RankT5 & .3866 & .1825 & .2774 & .8572 & .1972 & .3702 & .3511 & .0718 & .1289 & .8155 & .0751 & .1693  \\
    & \quad +LDIST  & .3894 & .1851 & .2787 & \underline{.8610} & .1996 & .3734 & .3690 & .0791 & .1366 & \underline{.8190} & .0825 & .1767  \\
    & \quad +LiT5   & .3915 & \underline{.1906} & \underline{.2842} & .8567 & \underline{.2053} & \underline{.3776} & .3676 & .0800 & .1369 & \underline{.8190} & .0833 & .1770 \\
    \cmidrule{2-14}
    & BGM          & .3723   & .1686 & .2723 & .8537 & .1824 & .3651 & .3296 & .0642 & .1282 & .8104 & .0677 & .1666  \\
    & \ours{} & \textbf{.4174 } & \textbf{.2247} & \textbf{.3055} & \textbf{.8637} & \textbf{.2392} & \textbf{.4015} & \textbf{.3951} & \textbf{.0942} & \textbf{.1456} & \textbf{.8256} & \textbf{.0976} & \textbf{.1863} \\
    \bottomrule
    \end{tabular}
    }
    \caption{
    Overall results of all models. The best and second-best results are in bold and underlined, respectively. 
    }
    \label{tab:overall}
\end{table*}

\section{Experiment}
\subsection{Datasets and Metrics}
\noindent \textit{Datasets}.
We conduct our experiments on two publicly available datasets that focus on multi-document summarization and long-form query-answer (QA) respectively, \ie, \textbf{WikiPassageQA}~\cite{wikiasp} and \textbf{WikiAsp}~\cite{wikiasp}. WikiPassageQA offers human-annotated quality-evaluated questions and long-form answers. We chose this dataset because its answers are generally comprehensive, related to various aspects of questions, and the answer length is fairly long. WikiAsp dataset is devised for generating aspect-based summaries of entities from 20 domains. We follow~\cite{flare} to convert it into open-domain QA settings. 
To support our experiments, we first operate some data pre-processing to ensure that each piece of data contains the question, ground truth answer, sub-aspects of the question, and their sub-answers. The process details and statistical information of datasets are demonstrated in Appendix~\ref{app:dataprocess}. 
\noindent\textit{Metrics}.
To measure the matching scores of models' responses with long-form ground truth answers, we select F1, Rouge(-2 and -L), and BERT-Score as evaluation metrics.
Furthermore, we leverage the com-rouge score, which is introduced in Eq.~\ref{eq:reward}, to assess the coverage of responses on sub-answers. We implement $\phi()$ in Eq.~\ref{eq:reward} by Rouge-2, and -L to build Com-Rouge-2 and -L evaluation metrics. For briefness, F1, R2, RL, BS, CR2, and CRL are utilized to represent these metrics. Following~\cite{gpt4-eval}, we also leverage GPT-4 to conduct a pairwise evaluation of our method with baseline models to confirm its effectiveness further. To comprehensively evaluate the effectiveness of our proposed ranking algorithm, we also provide the ranking performance comparison and analysis in Appendix~\ref{app:ranking}


\begin{figure*}[t]
    \centering
    \includegraphics[width=0.9\linewidth]{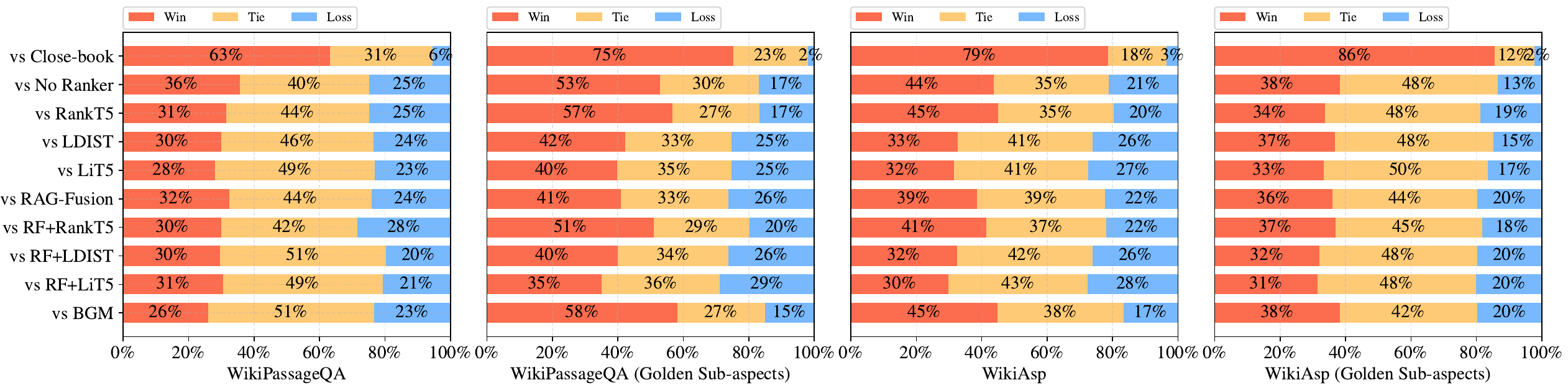}
    \caption{Results of the GPT-4-based evaluation comparing our method with baseline models.}
    \label{fig:gpt4-eval}
    \vspace{-10pt}
\end{figure*}
\subsection{Baselines}
To evaluate the effectiveness of our framework, we first build baselines with different RAG framework settings: (1) \textbf{Close-book} setting without external reference support. (2) ``Retrieve-Generation'' setting without ranking stage, namely \textbf{No Ranker}. (3) ``Retrieve-Rerank-Generation'' setting with various ranking algorithms, including \textbf{RankT5}~\cite{rankt5-orig,rankt5}, a point-wise T5-based ranking model, \textbf{LDist}~\cite{atlas}, a widely-used ranking algorithm in RAG systems~\cite{atlas,replug}, and \textbf{LiT5}~\cite{lit5}, a list-wise ranking model using slide-window-based ranking strategy. These baselines directly retrieve external documents based on the original queries. 
\textbf{RAG-Fusion}~\cite{ragfusion} proposes retrieving documents from various sub-aspects and providing the final ranking lists via a simple reciprocal rank fusion algorithm~\cite{RRF}. We set this framework as a basic baseline to compare the superiority of our proposed framework when explicitly considering query-aspects. We combine it with the above ranking algorithms to build various advanced versions of RAG-Fusion, \eg, \textbf{RAG-Fusion+RankT5}, etc. \textbf{BGM}~\cite{bgm} is a recent RAG framework that introduces PPO strategy to fine-tune list-wise ranking model based on the LLM's feedback. Due to limited space, we describe the implementation details in Appendix~\ref{app:implement}.

\subsection{Overall Results}
To fully evaluate the effectiveness of our proposed framework, \ours{}, we utilize two settings to conduct experiments, the first one we provide the predicted sub-aspects to retriever and ranker while in the second one, we provide the golden sub-aspects to unlock the \ours{}'s powers in the fullest extent possible. 
We present the overall results in Table~\ref{tab:overall} and Figure~\ref{fig:gpt4-eval}, and have the following conclusions:

(1) Whether given predicted or golden sub-aspects, \ours{} shows the best performance. This phenomenon confirms the ability of our framework to explore and leverage user's sub-intents underlying the issued multi-faceted questions, hence providing comprehensive responses. However, existing RAG systems solely consider query-document relevance without relationships among candidates, blocking their potential to understand user's various sub-intents and limiting the richness of final responses. 

(2) Compared to list-wise ranking algorithms, our method still illustrates better performance. Though there exist several list-wise ranking algorithms in the RAG community, such as LiT5 and BGM, these algorithms do not explicitly model the interactions among candidates from the perspective of comprehensiveness of the user intent coverage. Without such explicit guidance, these algorithms may be trapped in a locally optimal solution, hence impacting the overall quality of ranking lists.

(3) With golden sub-aspects, RAG-Fusion settings often outperform settings only considering the original question's retrieved documents. 
It reveals the importance of modeling the questions' sub-aspects in RAG systems for generating rich and reliable responses. However, with predicted sub-aspects, the RAG-Fusion variants do not outperform corresponding baselines without RAG-Fusion. This may be due to the gap between predicted sub-aspects and annotated sub-aspects. However, in real applications, the user's sub-intents may be diverse while we can only label some of them in datasets to evaluate model performances. Therefore, how to deal with such a gap between realistic and human annotation is still an open problem and will be further investigated in our future study.


\subsection{Ablation Studies}
\begin{figure}[!ht]
    \centering
    \includegraphics[width=1\linewidth]{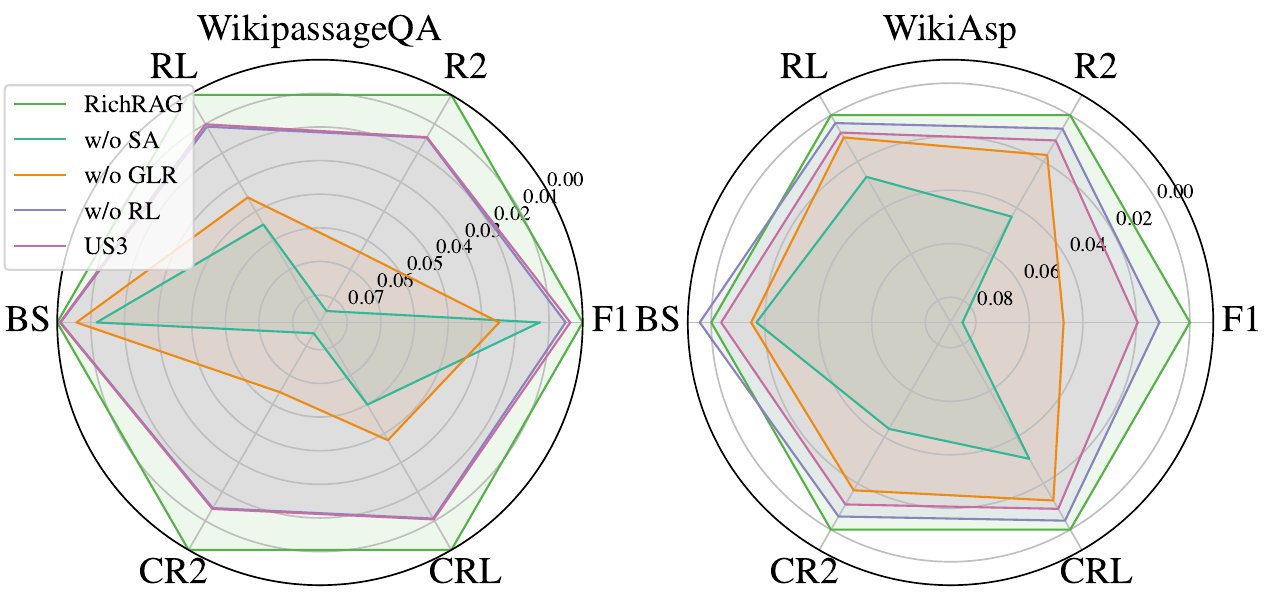}
    \caption{Ablation results of \ours{} on two datasets. }
    \label{fig:abl}
\end{figure}

In order to evaluate the role of our key modules, we further conduct the following ablation studies with results presented in Figure~\ref{fig:abl}. This figure shows the decline degrees of the ablation models compared with the complete \ours{}. 

(1) To confirm the importance of explicitly consideration of user's sub-intents, \ie question's sub-aspects. We further construct a variant, w/o SA, by directly ranking the retrieved documents of the original question without considering the candidate pool, $\mathcal{P}$. The significantly worse results than \ours{} further prove the importance of explicitly considering the various sub-intents underlying multi-faceted questions, which is beneficial for providing comprehensive responses for users.

(2) In our study, we propose a generative list-wise ranking module to generate LLM-preferred comprehensive ranking lists. To prove its advantages, we replace it with another list-wise algorithm, LiT5, in our framework to build a variant, w/o GLR. We find that it still underperforms \ours{}. This result validates the advantages of our model structure, \ie the ability to potentially model global interactions among various candidates with sub-aspects. While the sliding-window-based list-wise algorithm still has defects on it, hence limiting the performance.

(3) To confirm the role of alignment with LLMs' preferences, we build a variant of our framework, w/o. RL, which only supervised fine-tunes our ranker without the RL optimization. The declined performance proves that the LLMs' preferences are different from humans' preferences. As a result, it is vital for RAG systems to align the LLMs' preferences to enhance the overall quality of the final responses generated from LLMs. 

(4) To ensure the robustness and quality of the DPO algorithm, we propose a US$^3$ approach to build the pairwise training samples for it. To confirm its effect, we replace it by randomly creating the training pairs for the RL stage, building a variant, namely w/o US$^3$. The worse result than \ours{} proves the usefulness of this strategy. It suggests that the US$^3$ approach can create more reliable training pairs by ensuring the meaningful comparison between predictions of our ranker, hence optimizing it in a promising direction.


\subsection{Efficiency Analysis of Ranking Algorithms}
\begin{figure}[t]
    \centering
    \includegraphics[width=0.49\textwidth]{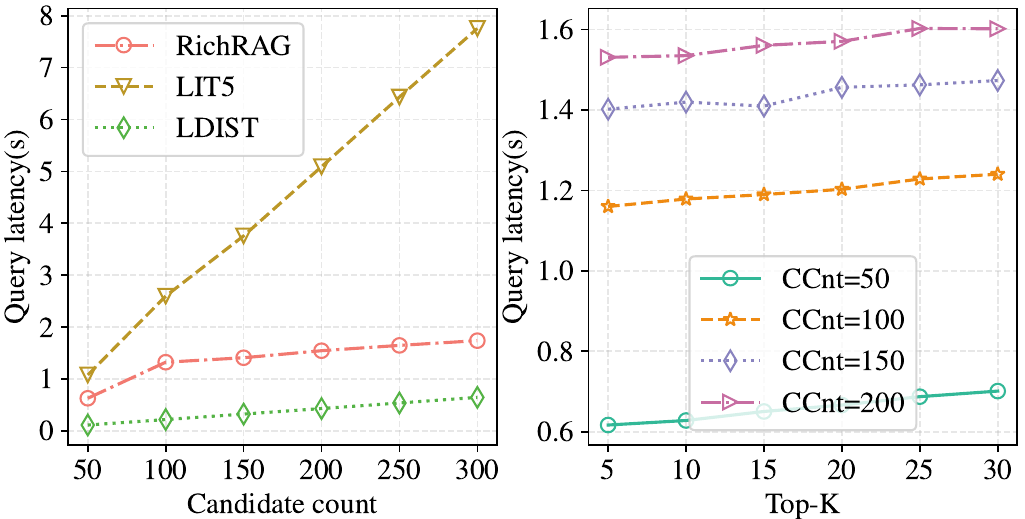}
    \caption{Efficiency experiments of different models.}
    \label{fig:efficient}
    \vspace{-10pt}
\end{figure}

We previously confirmed the advantages of considering various sub-aspects in RAG systems. However, with extensive candidate documents, the efficiency of ranking modules is also important. Therefore, we compare the query latency of our ranker to point-wise and list-wise ranking algorithms, LDIST and LiT5, to test their efficiency. First, we demonstrate their changes in query latency with the candidate amount in Figure~\ref{fig:efficient} (a). Obviously, our ranker has comparable efficiency and trend with the point-wise ranking algorithm. However, the time overhead of LiT5 rises more sharply along with the improvement of the candidate amount. This phenomenon proves that our ranker could provide a better trade-off between effectiveness and efficiency. Furthermore, since our ranker generates docids step-by-step, we further provide the trends of query latency with different generated document numbers and show the trend lines of different amounts of candidate documents in Figure~\ref{fig:efficient} (b). 
It can be found that as the number of ranked documents increases, all trendlines rise slowly, and the gap between different candidate counts (CCnt) is limited to 1 second. This phenomenon further proves the robustness of our ranker's efficiency across diverse ranking settings. 

Due to limited space, we provide further analysis studies in Appendix~\ref{app:ranking}, ~\ref{app:subcnt}, ~\ref{app:repetition}, ~\ref{app:topk} and ~\ref{app:case}.

\section{Conclusion}
In this study, we proposed a new RAG framework, \ours{}, to comprehensively consider the various sub-intents underlying users' broad questions, hence providing all-sided long-form responses for users. Specifically, we introduced a sub-aspect explorer to predict the potential sub-aspects contained by questions representing the user's sub-intents. According to sub-aspects and a fixed retriever, we could build extensive and diverse candidate pools. To provide comprehensive and LLM-preferred ranking lists, we designed a generative list-wise ranking model. It effectively and efficiently encodes the global relationships between candidates and multi-aspects, thereby offering global optimal ranking lists to LLMs. To ensure the ranking quality, we utilized a two-stage training process involving supervised fine-tuning and RL optimization. Furthermore, we devised a US$^3$ approach to create useful and reliable training samples to ensure the effectiveness of the DPO algorithm. Extensive experiments on two public datasets confirm the effectiveness and efficiency of \ours{}. 


\section*{Limitations}
In this work, we identified an underexplored but important scenario of RAG systems, where multi-faceted questions require rich and comprehensive responses satisfying various related sub-aspects. To handle these situations, we developed a framework, namely \ours{} to equip RAG models with the ability to generate rich and satisfying responses for multi-faceted questions. We acknowledge the following limitations of our current study that present opportunities for future investigations.

First, though we built an aspect explorer to identify users' sub-intents underlying multi-faceted questions, it is still shallow and there is a gap between predicted sub-aspects and real intents. This is because the user's intents are usually diverse and vary from person to person. Even though we annotated some sub-aspects in data samples, these may only cover a sub-set. Therefore, in this study, we mainly focus on how to provide a promising reference permutation given the user's potential sub-intents for enhancing the final generation to comprehensively respond to these sub-intents. 
Second, since the situation is still underexplored, few suitable datasets are available. The datasets we used in our study were chosen by carefully investigating the data samples' content and converted by some operations to a suitable data format. Therefore, the diversity of experiment datasets is limited. In the future, we will pay more attention to the evaluation and annotation of user intent exploration in such scenarios to support further study.



\bibliography{custom}

\begin{thebibliography}{44}
\providecommand{\natexlab}[1]{#1}

\bibitem[{Arora et~al.(2023)Arora, Kini, Chowdhury, Natarajan, Sinha, and Sharma}]{realm}
Daman Arora, Anush Kini, Sayak~Ray Chowdhury, Nagarajan Natarajan, Gaurav Sinha, and Amit Sharma. 2023.
\newblock \href {https://api.semanticscholar.org/CorpusID:264817661} {Gar-meets-rag paradigm for zero-shot information retrieval}.
\newblock \emph{ArXiv}, abs/2310.20158.

\bibitem[{Asai et~al.(2024)Asai, Wu, Wang, Sil, and Hajishirzi}]{self-rag}
Akari Asai, Zeqiu Wu, Yizhong Wang, Avirup Sil, and Hannaneh Hajishirzi. 2024.
\newblock \href {https://openreview.net/forum?id=hSyW5go0v8} {Self-{RAG}: Learning to retrieve, generate, and critique through self-reflection}.
\newblock In \emph{The Twelfth International Conference on Learning Representations}.

\bibitem[{Borgeaud et~al.(2022)Borgeaud, Mensch, Hoffmann, Cai, Rutherford, Millican, Van Den~Driessche, Lespiau, Damoc, Clark, De~Las~Casas, Guy, Menick, Ring, Hennigan, Huang, Maggiore, Jones, Cassirer, Brock, Paganini, Irving, Vinyals, Osindero, Simonyan, Rae, Elsen, and Sifre}]{retro}
Sebastian Borgeaud, Arthur Mensch, Jordan Hoffmann, Trevor Cai, Eliza Rutherford, Katie Millican, George~Bm Van Den~Driessche, Jean-Baptiste Lespiau, Bogdan Damoc, Aidan Clark, Diego De~Las~Casas, Aurelia Guy, Jacob Menick, Roman Ring, Tom Hennigan, Saffron Huang, Loren Maggiore, Chris Jones, Albin Cassirer, Andy Brock, Michela Paganini, Geoffrey Irving, Oriol Vinyals, Simon Osindero, Karen Simonyan, Jack Rae, Erich Elsen, and Laurent Sifre. 2022.
\newblock \href {https://proceedings.mlr.press/v162/borgeaud22a.html} {Improving language models by retrieving from trillions of tokens}.
\newblock In \emph{Proceedings of the 39th International Conference on Machine Learning}, volume 162 of \emph{Proceedings of Machine Learning Research}, pages 2206--2240. PMLR.

\bibitem[{Chan et~al.(2024)Chan, Xu, Yuan, Luo, Xue, Guo, and Fu}]{chan2024rqrag}
Chi-Min Chan, Chunpu Xu, Ruibin Yuan, Hongyin Luo, Wei Xue, Yike Guo, and Jie Fu. 2024.
\newblock \href {https://arxiv.org/abs/2404.00610} {Rq-rag: Learning to refine queries for retrieval augmented generation}.
\newblock \emph{Preprint}, arXiv:2404.00610.

\bibitem[{Chung et~al.(2022)Chung, Hou, Longpre, Zoph, Tay, Fedus, Li, Wang, Dehghani, Brahma, Webson, Gu, Dai, Suzgun, Chen, Chowdhery, Narang, Mishra, Yu, Zhao, Huang, Dai, Yu, Petrov, Chi, Dean, Devlin, Roberts, Zhou, Le, and Wei}]{flant5}
Hyung~Won Chung, Le~Hou, Shayne Longpre, Barret Zoph, Yi~Tay, William Fedus, Eric Li, Xuezhi Wang, Mostafa Dehghani, Siddhartha Brahma, Albert Webson, Shixiang~Shane Gu, Zhuyun Dai, Mirac Suzgun, Xinyun Chen, Aakanksha Chowdhery, Sharan Narang, Gaurav Mishra, Adams Yu, Vincent Zhao, Yanping Huang, Andrew Dai, Hongkun Yu, Slav Petrov, Ed~H. Chi, Jeff Dean, Jacob Devlin, Adam Roberts, Denny Zhou, Quoc~V. Le, and Jason Wei. 2022.
\newblock \href {https://doi.org/10.48550/ARXIV.2210.11416} {Scaling instruction-finetuned language models}.
\newblock \emph{arXiv preprint}.

\bibitem[{Cormack et~al.(2009)Cormack, Clarke, and Buettcher}]{RRF}
Gordon~V. Cormack, Charles L~A Clarke, and Stefan Buettcher. 2009.
\newblock \href {https://doi.org/10.1145/1571941.1572114} {Reciprocal rank fusion outperforms condorcet and individual rank learning methods}.
\newblock In \emph{Proceedings of the 32nd International ACM SIGIR Conference on Research and Development in Information Retrieval}, SIGIR '09, page 758–759, New York, NY, USA. Association for Computing Machinery.

\bibitem[{Dang and Croft(2012)}]{pm2}
Van Dang and W.~Bruce Croft. 2012.
\newblock \href {https://doi.org/10.1145/2348283.2348296} {Diversity by proportionality: an election-based approach to search result diversification}.
\newblock In \emph{Proceedings of the 35th International ACM SIGIR Conference on Research and Development in Information Retrieval}, SIGIR '12, page 65–74, New York, NY, USA. Association for Computing Machinery.

\bibitem[{Gao et~al.(2024)Gao, Xiong, Gao, Jia, Pan, Bi, Dai, Sun, Wang, and Wang}]{ragsurvey}
Yunfan Gao, Yun Xiong, Xinyu Gao, Kangxiang Jia, Jinliu Pan, Yuxi Bi, Yi~Dai, Jiawei Sun, Meng Wang, and Haofen Wang. 2024.
\newblock \href {https://arxiv.org/abs/2312.10997} {Retrieval-augmented generation for large language models: A survey}.
\newblock \emph{Preprint}, arXiv:2312.10997.

\bibitem[{Hayashi et~al.(2021)Hayashi, Budania, Wang, Ackerson, Neervannan, and Neubig}]{wikiasp}
Hiroaki Hayashi, Prashant Budania, Peng Wang, Chris Ackerson, Raj Neervannan, and Graham Neubig. 2021.
\newblock \href {https://doi.org/10.1162/tacl_a_00362} {{W}iki{A}sp: A dataset for multi-domain aspect-based summarization}.
\newblock \emph{Transactions of the Association for Computational Linguistics}, 9:211--225.

\bibitem[{Izacard and Grave(2021)}]{fid}
Gautier Izacard and Edouard Grave. 2021.
\newblock \href {https://doi.org/10.18653/v1/2021.eacl-main.74} {Leveraging passage retrieval with generative models for open domain question answering}.
\newblock In \emph{Proceedings of the 16th Conference of the European Chapter of the Association for Computational Linguistics: Main Volume}, pages 874--880, Online. Association for Computational Linguistics.

\bibitem[{Izacard et~al.(2024)Izacard, Lewis, Lomeli, Hosseini, Petroni, Schick, Dwivedi-Yu, Joulin, Riedel, and Grave}]{atlas}
Gautier Izacard, Patrick Lewis, Maria Lomeli, Lucas Hosseini, Fabio Petroni, Timo Schick, Jane Dwivedi-Yu, Armand Joulin, Sebastian Riedel, and Edouard Grave. 2024.
\newblock Atlas: few-shot learning with retrieval augmented language models.
\newblock \emph{J. Mach. Learn. Res.}, 24(1).

\bibitem[{Jiang et~al.(2023)Jiang, Xu, Gao, Sun, Liu, Dwivedi-Yu, Yang, Callan, and Neubig}]{flare}
Zhengbao Jiang, Frank Xu, Luyu Gao, Zhiqing Sun, Qian Liu, Jane Dwivedi-Yu, Yiming Yang, Jamie Callan, and Graham Neubig. 2023.
\newblock \href {https://doi.org/10.18653/v1/2023.emnlp-main.495} {Active retrieval augmented generation}.
\newblock In \emph{Proceedings of the 2023 Conference on Empirical Methods in Natural Language Processing}, pages 7969--7992, Singapore. Association for Computational Linguistics.

\bibitem[{Ke et~al.(2024)Ke, Kong, Li, Zhang, Mei, and Bendersky}]{bgm}
Zixuan Ke, Weize Kong, Cheng Li, Mingyang Zhang, Qiaozhu Mei, and Michael Bendersky. 2024.
\newblock \href {https://arxiv.org/abs/2401.06954} {Bridging the preference gap between retrievers and llms}.
\newblock \emph{Preprint}, arXiv:2401.06954.

\bibitem[{Khot et~al.(2023)Khot, Trivedi, Finlayson, Fu, Richardson, Clark, and Sabharwal}]{khot2023decomposed}
Tushar Khot, Harsh Trivedi, Matthew Finlayson, Yao Fu, Kyle Richardson, Peter Clark, and Ashish Sabharwal. 2023.
\newblock \href {https://arxiv.org/abs/2210.02406} {Decomposed prompting: A modular approach for solving complex tasks}.
\newblock \emph{Preprint}, arXiv:2210.02406.

\bibitem[{Lewis et~al.(2020)Lewis, Perez, Piktus, Petroni, Karpukhin, Goyal, K\"{u}ttler, Lewis, Yih, Rockt\"{a}schel, Riedel, and Kiela}]{rag}
Patrick Lewis, Ethan Perez, Aleksandra Piktus, Fabio Petroni, Vladimir Karpukhin, Naman Goyal, Heinrich K\"{u}ttler, Mike Lewis, Wen-tau Yih, Tim Rockt\"{a}schel, Sebastian Riedel, and Douwe Kiela. 2020.
\newblock Retrieval-augmented generation for knowledge-intensive nlp tasks.
\newblock In \emph{Proceedings of the 34th International Conference on Neural Information Processing Systems}, NIPS '20, Red Hook, NY, USA. Curran Associates Inc.

\bibitem[{Li et~al.(2024)Li, Liu, Xiong, Yu, Yan, Wang, and Yu}]{xiong_filter}
Xinze Li, Zhenghao Liu, Chenyan Xiong, Shi Yu, Yukun Yan, Shuo Wang, and Ge~Yu. 2024.
\newblock \href {https://arxiv.org/abs/2402.16058} {Say more with less: Understanding prompt learning behaviors through gist compression}.
\newblock \emph{Preprint}, arXiv:2402.16058.

\bibitem[{Lin et~al.(2023)Lin, Chen, Chen, Shi, Lomeli, James, Rodriguez, Kahn, Szilvasy, Lewis, Zettlemoyer, and Yih}]{radit}
Xi~Victoria Lin, Xilun Chen, Mingda Chen, Weijia Shi, Maria Lomeli, Rich James, Pedro Rodriguez, Jacob Kahn, Gergely Szilvasy, Mike Lewis, Luke Zettlemoyer, and Scott Yih. 2023.
\newblock \href {https://api.semanticscholar.org/CorpusID:263605962} {Ra-dit: Retrieval-augmented dual instruction tuning}.
\newblock \emph{ArXiv}, abs/2310.01352.

\bibitem[{Liu et~al.(2020)Liu, Dou, Wang, Lu, and Wen}]{DVGAN}
Jiongnan Liu, Zhicheng Dou, Xiaojie Wang, Shuqi Lu, and Ji-Rong Wen. 2020.
\newblock \href {https://doi.org/10.1145/3397271.3401084} {Dvgan: A minimax game for search result diversification combining explicit and implicit features}.
\newblock In \emph{Proceedings of the 43rd International ACM SIGIR Conference on Research and Development in Information Retrieval}, SIGIR '20, page 479–488, New York, NY, USA. Association for Computing Machinery.

\bibitem[{Loshchilov and Hutter(2019)}]{ICLR19_AdamW}
Ilya Loshchilov and Frank Hutter. 2019.
\newblock \href {https://openreview.net/forum?id=Bkg6RiCqY7} {Decoupled weight decay regularization}.
\newblock In \emph{7th International Conference on Learning Representations, {ICLR} 2019, New Orleans, LA, USA, May 6-9, 2019}.

\bibitem[{Nogueira et~al.(2020)Nogueira, Jiang, Pradeep, and Lin}]{rankt5-orig}
Rodrigo Nogueira, Zhiying Jiang, Ronak Pradeep, and Jimmy Lin. 2020.
\newblock \href {https://doi.org/10.18653/v1/2020.findings-emnlp.63} {Document ranking with a pretrained sequence-to-sequence model}.
\newblock In \emph{Findings of the Association for Computational Linguistics: EMNLP 2020}, pages 708--718, Online. Association for Computational Linguistics.

\bibitem[{Paranjape et~al.(2022)Paranjape, Khattab, Potts, Zaharia, and Manning}]{hindsight}
Ashwin Paranjape, Omar Khattab, Christopher Potts, Matei Zaharia, and Christopher~D. Manning. 2022.
\newblock \href {https://openreview.net/forum?id=Vr\_BTpw3wz} {Hindsight: Posterior-guided training of retrievers for improved open-ended generation}.
\newblock In \emph{The Tenth International Conference on Learning Representations, {ICLR} 2022, Virtual Event, April 25-29, 2022}. OpenReview.net.

\bibitem[{Qin et~al.(2024)Qin, Jagerman, Hui, Zhuang, Wu, Yan, Shen, Liu, Liu, Metzler, Wang, and Bendersky}]{pair2}
Zhen Qin, Rolf Jagerman, Kai Hui, Honglei Zhuang, Junru Wu, Le~Yan, Jiaming Shen, Tianqi Liu, Jialu Liu, Donald Metzler, Xuanhui Wang, and Michael Bendersky. 2024.
\newblock \href {https://arxiv.org/abs/2306.17563} {Large language models are effective text rankers with pairwise ranking prompting}.
\newblock \emph{Preprint}, arXiv:2306.17563.

\bibitem[{Rackauckas(2024)}]{ragfusion}
Zackary Rackauckas. 2024.
\newblock \href {https://doi.org/10.5121/ijnlc.2024.13103} {Rag-fusion: A new take on retrieval augmented generation}.
\newblock \emph{International Journal on Natural Language Computing}, 13(1):37–47.

\bibitem[{Rafailov et~al.(2023)Rafailov, Sharma, Mitchell, Manning, Ermon, and Finn}]{DPO}
Rafael Rafailov, Archit Sharma, Eric Mitchell, Christopher~D Manning, Stefano Ermon, and Chelsea Finn. 2023.
\newblock \href {https://openreview.net/forum?id=HPuSIXJaa9} {Direct preference optimization: Your language model is secretly a reward model}.
\newblock In \emph{Thirty-seventh Conference on Neural Information Processing Systems}.

\bibitem[{Raffel et~al.(2020)Raffel, Shazeer, Roberts, Lee, Narang, Matena, Zhou, Li, and Liu}]{t5}
Colin Raffel, Noam Shazeer, Adam Roberts, Katherine Lee, Sharan Narang, Michael Matena, Yanqi Zhou, Wei Li, and Peter~J. Liu. 2020.
\newblock \href {http://jmlr.org/papers/v21/20-074.html} {Exploring the limits of transfer learning with a unified text-to-text transformer}.
\newblock \emph{Journal of Machine Learning Research}, 21(140):1--67.

\bibitem[{Santos et~al.(2010)Santos, Macdonald, and Ounis}]{xquad}
Rodrygo~L.T. Santos, Craig Macdonald, and Iadh Ounis. 2010.
\newblock \href {https://doi.org/10.1145/1772690.1772780} {Exploiting query reformulations for web search result diversification}.
\newblock In \emph{Proceedings of the 19th International Conference on World Wide Web}, WWW '10, page 881–890, New York, NY, USA. Association for Computing Machinery.

\bibitem[{Shi et~al.(2023)Shi, Min, Yasunaga, Seo, James, Lewis, Zettlemoyer, and tau Yih}]{replug}
Weijia Shi, Sewon Min, Michihiro Yasunaga, Minjoon Seo, Rich James, Mike Lewis, Luke Zettlemoyer, and Wen tau Yih. 2023.
\newblock \href {https://arxiv.org/abs/2301.12652} {Replug: Retrieval-augmented black-box language models}.
\newblock \emph{Preprint}, arXiv:2301.12652.

\bibitem[{Sun et~al.(2023)Sun, Yan, Ma, Wang, Ren, Chen, Yin, and Ren}]{rankgpt}
Weiwei Sun, Lingyong Yan, Xinyu Ma, Shuaiqiang Wang, Pengjie Ren, Zhumin Chen, Dawei Yin, and Zhaochun Ren. 2023.
\newblock \href {https://doi.org/10.18653/v1/2023.emnlp-main.923} {Is {C}hat{GPT} good at search? investigating large language models as re-ranking agents}.
\newblock In \emph{Proceedings of the 2023 Conference on Empirical Methods in Natural Language Processing}, pages 14918--14937, Singapore. Association for Computational Linguistics.

\bibitem[{Tamber et~al.(2023)Tamber, Pradeep, and Lin}]{lit5}
Manveer~Singh Tamber, Ronak Pradeep, and Jimmy Lin. 2023.
\newblock \href {https://arxiv.org/abs/2312.16098} {Scaling down, litting up: Efficient zero-shot listwise reranking with seq2seq encoder-decoder models}.
\newblock \emph{Preprint}, arXiv:2312.16098.

\bibitem[{Touvron et~al.(2023)Touvron, Martin, Stone, Albert, Almahairi, Babaei, Bashlykov, Batra, Bhargava, Bhosale, Bikel, Blecher, Ferrer, Chen, Cucurull, Esiobu, Fernandes, Fu, Fu, Fuller, Gao, Goswami, Goyal, Hartshorn, Hosseini, Hou, Inan, Kardas, Kerkez, Khabsa, Kloumann, Korenev, Koura, Lachaux, Lavril, Lee, Liskovich, Lu, Mao, Martinet, Mihaylov, Mishra, Molybog, Nie, Poulton, Reizenstein, Rungta, Saladi, Schelten, Silva, Smith, Subramanian, Tan, Tang, Taylor, Williams, Kuan, Xu, Yan, Zarov, Zhang, Fan, Kambadur, Narang, Rodriguez, Stojnic, Edunov, and Scialom}]{llama}
Hugo Touvron, Louis Martin, Kevin Stone, Peter Albert, Amjad Almahairi, Yasmine Babaei, Nikolay Bashlykov, Soumya Batra, Prajjwal Bhargava, Shruti Bhosale, Dan Bikel, Lukas Blecher, Cristian~Canton Ferrer, Moya Chen, Guillem Cucurull, David Esiobu, Jude Fernandes, Jeremy Fu, Wenyin Fu, Brian Fuller, Cynthia Gao, Vedanuj Goswami, Naman Goyal, Anthony Hartshorn, Saghar Hosseini, Rui Hou, Hakan Inan, Marcin Kardas, Viktor Kerkez, Madian Khabsa, Isabel Kloumann, Artem Korenev, Punit~Singh Koura, Marie-Anne Lachaux, Thibaut Lavril, Jenya Lee, Diana Liskovich, Yinghai Lu, Yuning Mao, Xavier Martinet, Todor Mihaylov, Pushkar Mishra, Igor Molybog, Yixin Nie, Andrew Poulton, Jeremy Reizenstein, Rashi Rungta, Kalyan Saladi, Alan Schelten, Ruan Silva, Eric~Michael Smith, Ranjan Subramanian, Xiaoqing~Ellen Tan, Binh Tang, Ross Taylor, Adina Williams, Jian~Xiang Kuan, Puxin Xu, Zheng Yan, Iliyan Zarov, Yuchen Zhang, Angela Fan, Melanie Kambadur, Sharan Narang, Aurelien Rodriguez, Robert Stojnic, Sergey Edunov, and Thomas
  Scialom. 2023.
\newblock \href {https://arxiv.org/abs/2307.09288} {Llama 2: Open foundation and fine-tuned chat models}.
\newblock \emph{Preprint}, arXiv:2307.09288.

\bibitem[{Wang et~al.(2023{\natexlab{a}})Wang, Ping, Xu, McAfee, Liu, Shoeybi, Dong, Kuchaiev, Li, Xiao, Anandkumar, and Catanzaro}]{retroplus}
Boxin Wang, Wei Ping, Peng Xu, Lawrence McAfee, Zihan Liu, Mohammad Shoeybi, Yi~Dong, Oleksii Kuchaiev, Bo~Li, Chaowei Xiao, Anima Anandkumar, and Bryan Catanzaro. 2023{\natexlab{a}}.
\newblock \href {https://doi.org/10.18653/v1/2023.emnlp-main.482} {Shall we pretrain autoregressive language models with retrieval? a comprehensive study}.
\newblock In \emph{Proceedings of the 2023 Conference on Empirical Methods in Natural Language Processing}, pages 7763--7786, Singapore. Association for Computational Linguistics.

\bibitem[{Wang et~al.(2024{\natexlab{a}})Wang, Xue, Zhou, Zhang, Wang, Chen, Wang, and fai Wong}]{wang2024selfdc}
Hongru Wang, Boyang Xue, Baohang Zhou, Tianhua Zhang, Cunxiang Wang, Guanhua Chen, Huimin Wang, and Kam fai Wong. 2024{\natexlab{a}}.
\newblock \href {https://arxiv.org/abs/2402.13514} {Self-dc: When to retrieve and when to generate? self divide-and-conquer for compositional unknown questions}.
\newblock \emph{Preprint}, arXiv:2402.13514.

\bibitem[{Wang et~al.(2023{\natexlab{b}})Wang, Dou, Yao, Zhou, and Wen}]{expliPS}
Shuting Wang, Zhicheng Dou, Jing Yao, Yujia Zhou, and Ji-Rong Wen. 2023{\natexlab{b}}.
\newblock \href {https://doi.org/10.1145/3543507.3583488} {Incorporating explicit subtopics in personalized search}.
\newblock In \emph{Proceedings of the ACM Web Conference 2023}, WWW '23, page 3364–3374, New York, NY, USA. Association for Computing Machinery.

\bibitem[{Wang et~al.(2024{\natexlab{b}})Wang, Song, Cheng, Fu, Guo, Fang, Zhu, and Dou}]{wang2024domainrag}
Shuting Wang, Jiongnan Liu~Shiren Song, Jiehan Cheng, Yuqi Fu, Peidong Guo, Kun Fang, Yutao Zhu, and Zhicheng Dou. 2024{\natexlab{b}}.
\newblock \href {https://arxiv.org/abs/2406.05654} {Domainrag: A chinese benchmark for evaluating domain-specific retrieval-augmented generation}.
\newblock \emph{Preprint}, arXiv:2406.05654.

\bibitem[{Wang et~al.(2023{\natexlab{c}})Wang, Yu, Tan, O'Brien, Pasunuru, Dwivedi-Yu, Golovneva, Zettlemoyer, Fazel-Zarandi, and Celikyilmaz}]{gpt4-eval}
Tianlu Wang, Ping Yu, Xiaoqing~Ellen Tan, Sean O'Brien, Ramakanth Pasunuru, Jane Dwivedi-Yu, Olga Golovneva, Luke Zettlemoyer, Maryam Fazel-Zarandi, and Asli Celikyilmaz. 2023{\natexlab{c}}.
\newblock \href {https://arxiv.org/abs/2308.04592} {Shepherd: A critic for language model generation}.
\newblock \emph{Preprint}, arXiv:2308.04592.

\bibitem[{Wang et~al.(2023{\natexlab{d}})Wang, Araki, Jiang, Parvez, and Neubig}]{zhengbao_filter}
Zhiruo Wang, Jun Araki, Zhengbao Jiang, Md~Rizwan Parvez, and Graham Neubig. 2023{\natexlab{d}}.
\newblock \href {https://arxiv.org/abs/2311.08377} {Learning to filter context for retrieval-augmented generation}.
\newblock \emph{Preprint}, arXiv:2311.08377.

\bibitem[{Xiao et~al.(2023)Xiao, Liu, Zhang, and Muennighoff}]{bge_embedding}
Shitao Xiao, Zheng Liu, Peitian Zhang, and Niklas Muennighoff. 2023.
\newblock \href {https://arxiv.org/abs/2309.07597} {C-pack: Packaged resources to advance general chinese embedding}.
\newblock \emph{Preprint}, arXiv:2309.07597.

\bibitem[{Xu et~al.(2023)Xu, Shi, and Choi}]{comp1}
Fangyuan Xu, Weijia Shi, and Eunsol Choi. 2023.
\newblock \href {https://arxiv.org/abs/2310.04408} {Recomp: Improving retrieval-augmented lms with compression and selective augmentation}.
\newblock \emph{Preprint}, arXiv:2310.04408.

\bibitem[{Xu et~al.(2024{\natexlab{a}})Xu, Pang, Shen, Cheng, and Chua}]{searchchain}
Shicheng Xu, Liang Pang, Huawei Shen, Xueqi Cheng, and Tat-Seng Chua. 2024{\natexlab{a}}.
\newblock \href {https://arxiv.org/abs/2304.14732} {Search-in-the-chain: Interactively enhancing large language models with search for knowledge-intensive tasks}.
\newblock \emph{Preprint}, arXiv:2304.14732.

\bibitem[{Xu et~al.(2024{\natexlab{b}})Xu, Pang, Xu, Shen, and Cheng}]{xu2024listaware}
Shicheng Xu, Liang Pang, Jun Xu, Huawei Shen, and Xueqi Cheng. 2024{\natexlab{b}}.
\newblock \href {https://arxiv.org/abs/2402.02764} {List-aware reranking-truncation joint model for search and retrieval-augmented generation}.
\newblock \emph{Preprint}, arXiv:2402.02764.

\bibitem[{Yao et~al.(2023)Yao, Zhao, Yu, Du, Shafran, Narasimhan, and Cao}]{yao2023react}
Shunyu Yao, Jeffrey Zhao, Dian Yu, Nan Du, Izhak Shafran, Karthik Narasimhan, and Yuan Cao. 2023.
\newblock \href {https://arxiv.org/abs/2210.03629} {React: Synergizing reasoning and acting in language models}.
\newblock \emph{Preprint}, arXiv:2210.03629.

\bibitem[{Yu et~al.(2023)Yu, Xiong, Yu, and Liu}]{ARR}
Zichun Yu, Chenyan Xiong, Shi Yu, and Zhiyuan Liu. 2023.
\newblock \href {https://doi.org/10.18653/v1/2023.acl-long.136} {Augmentation-adapted retriever improves generalization of language models as generic plug-in}.
\newblock In \emph{Proceedings of the 61st Annual Meeting of the Association for Computational Linguistics (Volume 1: Long Papers)}, pages 2421--2436, Toronto, Canada. Association for Computational Linguistics.

\bibitem[{Zhuang et~al.(2023{\natexlab{a}})Zhuang, Qin, Jagerman, Hui, Ma, Lu, Ni, Wang, and Bendersky}]{rankt5}
Honglei Zhuang, Zhen Qin, Rolf Jagerman, Kai Hui, Ji~Ma, Jing Lu, Jianmo Ni, Xuanhui Wang, and Michael Bendersky. 2023{\natexlab{a}}.
\newblock \href {https://doi.org/10.1145/3539618.3592047} {Rankt5: Fine-tuning t5 for text ranking with ranking losses}.
\newblock In \emph{Proceedings of the 46th International ACM SIGIR Conference on Research and Development in Information Retrieval}, SIGIR '23, page 2308–2313, New York, NY, USA. Association for Computing Machinery.

\bibitem[{Zhuang et~al.(2023{\natexlab{b}})Zhuang, Zhuang, Koopman, and Zuccon}]{pair1}
Shengyao Zhuang, Honglei Zhuang, Bevan Koopman, and Guido Zuccon. 2023{\natexlab{b}}.
\newblock \href {https://arxiv.org/abs/2310.09497} {A setwise approach for effective and highly efficient zero-shot ranking with large language models}.
\newblock \emph{Preprint}, arXiv:2310.09497.

\end{thebibliography}
\appendix


\section{Data Pre-process}
\label{app:dataprocess}
Specifically, for the WikiPassageQA dataset, we prompt GPT-4 to generate appropriate and accurate sub-aspects and sub-answers for each data point to support our study, where sub-answers are split from the original long-form answers and sub-aspects are closely related to original questions and sub-answers. The content of the prompt is shown in Prompt~\ref{gpt}. 
To ensure the quality of reformulated WikiPassageQA, the prompt of GPT-4 is decided via the following steps:

(1) \textit{Human-Annotated Demonstrations}: We first selected five data examples and manually annotated their sub-aspects and sub-answers. These human-annotated examples are used as demonstrations to prompt the GPT-4 to generate satisfied results.

(2) \textit{Calibration with Human verification}: Then, we randomly sampled 50 examples and used them as calibration examples. Specifically, we ceaselessly adjusted the content of the prompt based on the generation quality of these calibration examples until we thought the generation quality of these examples was satisfied. The satisfaction rate of human evaluation was 94\%.

The WikiAsp dataset is devised for generating aspect-based summaries of entities from 20 domains. Each piece of data is built from a Wikipedia article, consisting of various aspects of this article and aspect-based summaries. Each target summary corresponds to an aspect. We follow~\cite{flare} to convert it into open-domain QA settings and introduce some additional operations to make it suitable for our experimental settings. Firstly, samples with more than two aspects are retained, and the remaining parts are removed to ensure the multifaceted nature of the experimental samples. Then, due to the expensive costs of experiments and the large amount of the whole dataset, we evenly sample subsets from each domain to build the experimental samples and split training, validation, and test sets according to the ratio of 10:1:1. Finally, we insert the title of the original Wikipedia article into a template: ``Generate a summary about {title}'' to mimic the real question format, hence constructing the question of each sample data. The aspects and aspect-based summaries are treated as sub-aspects of the question and sub-answers. We concatenate these sub-answers to build the long-form answer for each sample.

The statistical information of our datasets is presented in Table~\ref{tbl:statistic}.
\begin{table}[!ht]
\small
    \centering
    \begin{tabular}{lrrr}
    \toprule
    \multirow{2}*{Items} & \multicolumn{3}{c}{WikiPassageQA}  \\
    \cmidrule(lr){2-4}
    & Train & Validation & Test \\
    \midrule
    Count          & 3,311   & 415    & 416   \\
    Avg. Q Len     & 9.53   & 9.70   & 9.44  \\
    Avg. A Len     & 148.14 & 145.93 & 146.1 \\
    Avg. SubCnt    & 3.77   & 3.77   & 3.78  \\
    Avg. Sub Q Len & 6.34   & 6.25   & 6.29  \\
    Avg. Sub A Len & 62.84  & 62.32  & 61.99 \\
    \toprule
    \multirow{2}*{Items} & \multicolumn{3}{c}{WikiAsp} \\
    \cmidrule(lr){2-4}
    & Train & Validation & Test \\
    \midrule
    Count          & 8,613   & 859    & 867    \\
    Avg. Q Len     & 7.01   & 6.97   & 6.94   \\
    Avg. A Len     & 201.7  & 216.26 & 200.44 \\
    Avg. SubCnt    & 2.38   & 2.41   & 2.40   \\
    Avg. Sub Q Len & 1.28   & 1.28   & 1.29   \\
    Avg. Sub A Len & 229.36 & 240.6  & 221.91 \\
    \bottomrule
    \end{tabular}
    \caption{Statistical information of datasets. }
    \label{tbl:statistic}
\end{table}


\section{Implementation Details}\label{app:implement}
The sub-aspect explorer is implemented by Llama-2-7B-chat~\cite{llama}. We set the learning rate as 5e-5, batch size as 64, and use AdamW~\cite{ICLR19_AdamW} to fine-tune it. 
For the multi-faceted retriever, the number of retrieved documents for each sub-aspect is set as 50. The maximum capacity of the pool is 290 for WikipasssageQA and 270 for WikiAsp. 
For the generative list-wise ranker, We base on Flan-T5-base~\cite{flant5} to initialize it and rerank the top-10 final documents as provided external knowledge for the generator. In the SFT stage, we set the learning rate as 5e-5, batch size as 64, temperature $\tau$ as 0.1, and optimize the ranker to generate top-$10$ ranked document IDs with AdamW algorithm. In the RL stage, Llama-2-13B-chat is chosen as the generator, $\mathcal{G}$, providing reward feedback to the policy model. Then, we set $\mu$ as 0.1 to build 6,000 training pairs for the DPO algorithm. The batch size and learning rate are set as 32 and 3e-6 respectively to further optimize our ranker via the DPO strategy. 
We followed~\cite{atlas} to consider the Dec. 20, 2021, Wikipedia dump as our knowledge base and utilize BGE-en-base~\cite{bge_embedding} as our fixed retriever. All our baselines are optimized and evaluated by the same training, validation, and test datasets. Since the WikiAsp dataset has no relevance labels on documents, for each query,we view the documents whose matching score (the average of rouge scores evaluated by the ground truth answer) is higher than 0.5 as relevant documents to support the training of our baseline ranking models. Our experiments are conducted on the platform with four NVIDIA A100-SXM4-80GB GPUs. We will release our codes upon the acceptance of our study.

\begin{table*}[!t]
\small
    \centering
    
    \setlength{\tabcolsep}{1.6mm}{
    \begin{tabular}{lcccccccccccccc}
    \toprule
    \multirow{2}*{Models} & \multicolumn{6}{c}{WikiPassageQA} &  \multicolumn{6}{c}{WikiAsp} \\
    \cmidrule(r){2-7}\cmidrule(l){8-13}
    & {MAP} & {N1} & {N3} & {N5} & {N10} & {NCOM} & {MAP} & {N1} & {N3} & {N5} & {N10} & {NCOM}  \\
    \midrule
    Retriever      & .3201  & .2476 & .2933 & .3498 & .3951 & .7006 & .1780 & .1003 & .1678 & .1953 & .2294 & .3926 \\ 
    RankT5 (point) & .5200  & .4808 & .5176 & .5424 & .5916 & .7865 & .3712 & .3230 & .3704 & .3942 & .4043 & .4739 \\ 
    RankT5 (pair)  & .3602  & .2740 & .3367 & .3863 & .4298 & .7395 & .3299 & .2088 & .3274 & .3717 & .3894 & .5297   \\ 
    LDIST  (point) & \underline{.5346} & \underline{.5024} & \underline{.5261} & \underline{.5573} & \underline{.5979} & .7608 & \underline{.4051} & \underline{.3668} & \underline{.4078} & \underline{.4213} & \underline{.4280} & .5043 \\ 
    LiT5  (list)   & .3756  & .2813 & .3661 & .3968 & .4412 & .7465 & .2757 & .1696 & .2742 & .3142 & .3318 & \underline{.5442} \\ 
    \cmidrule{1-13} 
    RAG-Fusion            & .2632  & .1923 & .2348 & .2754 & .3317 & .5771 & .2040 & .1188 & .1844 & .2271 & .2608 & .4415 \\
    \quad +RankT5 (point) & .3962  & .3510 & .3831 & .4238 & .4641 & .6543 & .1591 & .0681 & .1411 & .1742 & .2177 & .4610 \\
    \quad +RankT5 (pair)  & .3946  & .3029 & .3669 & .4202 & .4758 & .6717 & .3297 & .2376 & .3249 & .3583 & .3869 & .4992 \\
    \quad +LDIST (point)  & .3746  & .3245 & .3654 & .3951 & .4385 & .6370 & .3345 & .2987 & .3241 & .3505 & .3733 & .5138 \\
    \quad +LiT5 (list)    & .3114  & .2308 & .2811 & .3244 & .3839 & .6547 & .2737 & .1995 & .2606 & .3014 & .3255 & .5196 \\
    \cmidrule{1-13}
    BGM         (list)  & .2614  & .2476 & .2480 & .2626 & .2864 & \underline{.7988} & .1382 & .0727 & .1152 & .1497 & .1921 & .3887 \\
    \ours{}     (list) & \textbf{.5444} & \textbf{.5240} & \textbf{.5359} & \textbf{.5663} & \textbf{.6035} & \textbf{.8065} & \textbf{.4880} & \textbf{.4556} & \textbf{.4962} & \textbf{.5047} & \textbf{.5067} & \textbf{.8935} \\
    \bottomrule
    \end{tabular}
    }
    \caption{
    Overall ranking results. The best and second-best results are in bold and underlined, respectively. 
    }
    \label{tab:ranking}
\end{table*}

\section{Ranking Performance}\label{app:ranking}
To prove the effectiveness of our proposed list-wise ranking module, we also evaluate the ranking modules' performance of our method and baselines. We indicate the types of ranking algorithms, including point-wise and list-wise. We further implement a pairwise version of the RankT5 model by using the following training objective to optimize it:
\begin{equation}
    \begin{aligned}
        \mathcal{L} = \!\!\!\!\!\!\!\!\!\!\!\!\!\!\sum_{d_1,d_2\in \mathcal{D}, R(d_1) > R(d_2)}\!\!\!\!\!\!\!\!\!\!\!\!\!\!\max(0, s(d_2) - s(d_1) + \gamma ), \;\; \gamma=1, \nonumber
    \end{aligned}
\end{equation}
where $\mathcal{D}$ denotes the candidate documents, $R(d)$ represents the relevance label of the document and $s(d)$ is the predicted score of the document. 

Specifically, we select some widely-used metrics, MAP and NDCG@k(k=1,3,5,10) to assess the ability to predict the document relevance of models. Furthermore, our model not only focuses on document relevance but also the comprehensiveness of the provided ranking lists. Thus, we propose a novel ranking metric, Normalized Comprehensiveness (NCOM), to assess the comprehensiveness of ranking lists, following the way to build the silver ranking targets (introduced in Section~\ref{sec:sft}). This metric considers both the document relevance and the coverage of ranking lists on sub-aspects, hence reliably evaluating the comprehensiveness of ranking lists. Its calculation is presented below:

Given a question $q$, its sub-answers $\{a_1,...a_n\}$, a pool of candidate document $\mathcal{P}$, a ranking list generated by a ranking model, $L=[d^1, ...,d^K]$, and a silver ranking list, $L^*$, generated by Eq. 8 in our paper, we calculate the generated and silver ranking lists' comprehensiveness scores as follow:
\begin{equation}
    \begin{aligned}
        COM(L) &= \sum_{t=1}^K\sum_{i=1}^n w^t_i\cdot \phi(d^t, a_i),\\
        w^t_i &=1-\mathrm{Norm}_i(\max_{d\in L[:t]}\phi(d,a_i))
    \end{aligned}
\end{equation}
where $w^t_i$ denotes the importance of $i$-th sub-aspects at the $t$-th step (which is the same as the Equation~\ref{eq:sub-weight} in our paper.), and $\phi(\cdot, \cdot)$ calculates the similarity between two sentences, \ie, average of Rouge-2, and Rouge-L scores.

The final normalized comprehensiveness score is computed by normalizing $COM(L)$ by $COM(L^*)$:
$$NCOM = COM(L) / COM(L^*).$$
The evaluation results are shown in Table~\ref{tbl:statistic}. From the experimental results, we can find that our method outperforms all types of ranking algorithms on both datasets and all metrics, especially on NCOM. This phenomenon proves that our ranking model could consider the document's relevance and list comprehensiveness simultaneously, hence providing comprehensive external knowledge and stimulating LLMs to generate better responses. We also notice that even though LiT5 and BGM are list-wise ranking models, their optimization does not explicitly consider the relationships among documents and, hence cannot provide satisfactory results. It also confirms the superiority of our proposed training algorithm.

\begin{figure*}[!ht]
	\centering
    \includegraphics[width=1\linewidth]{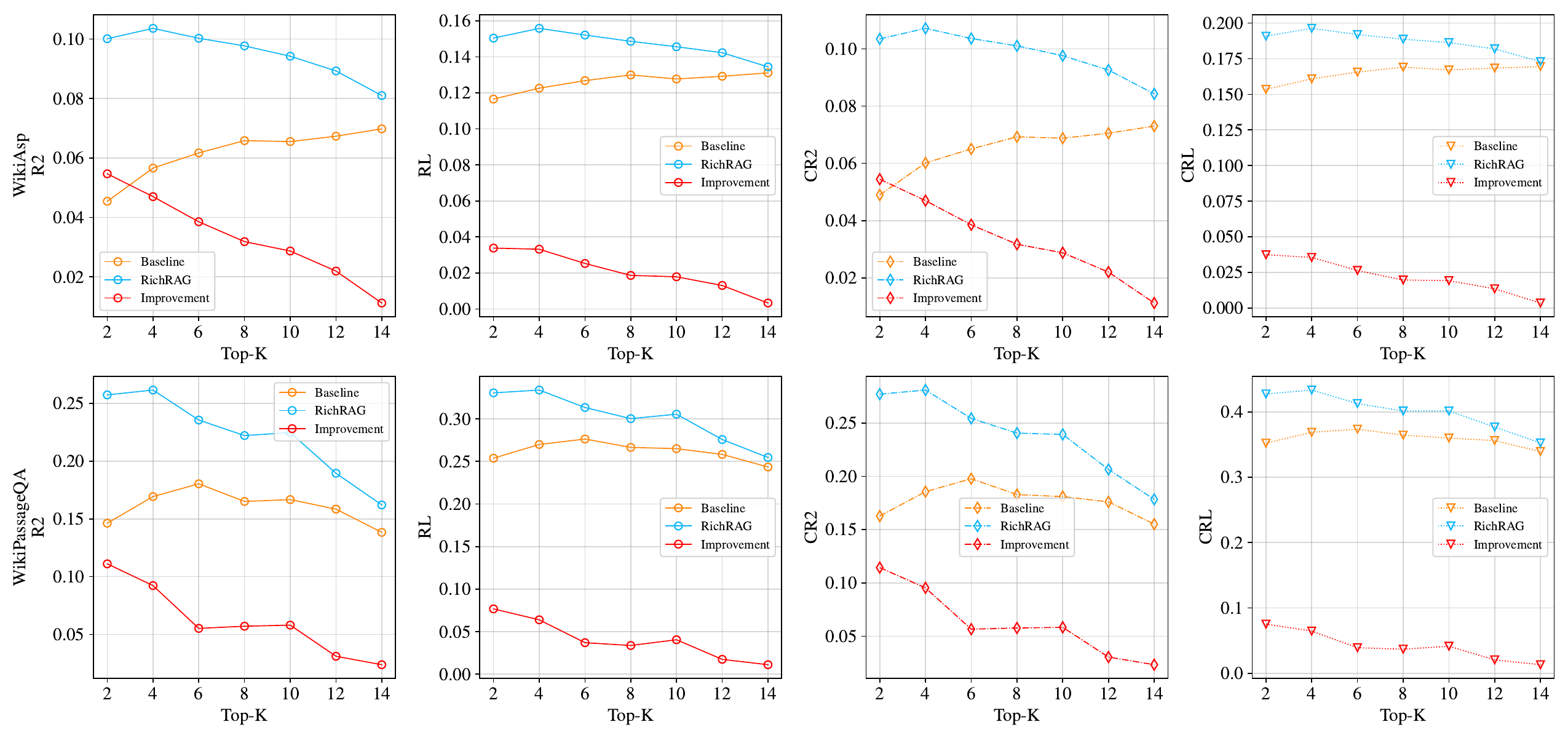}
    \caption{Trend of model performance as Top-K changes on both two dataset.}
    \label{fig:topk}
\end{figure*}

\begin{figure}[t]
    \centering
    \includegraphics[width=0.5\textwidth]{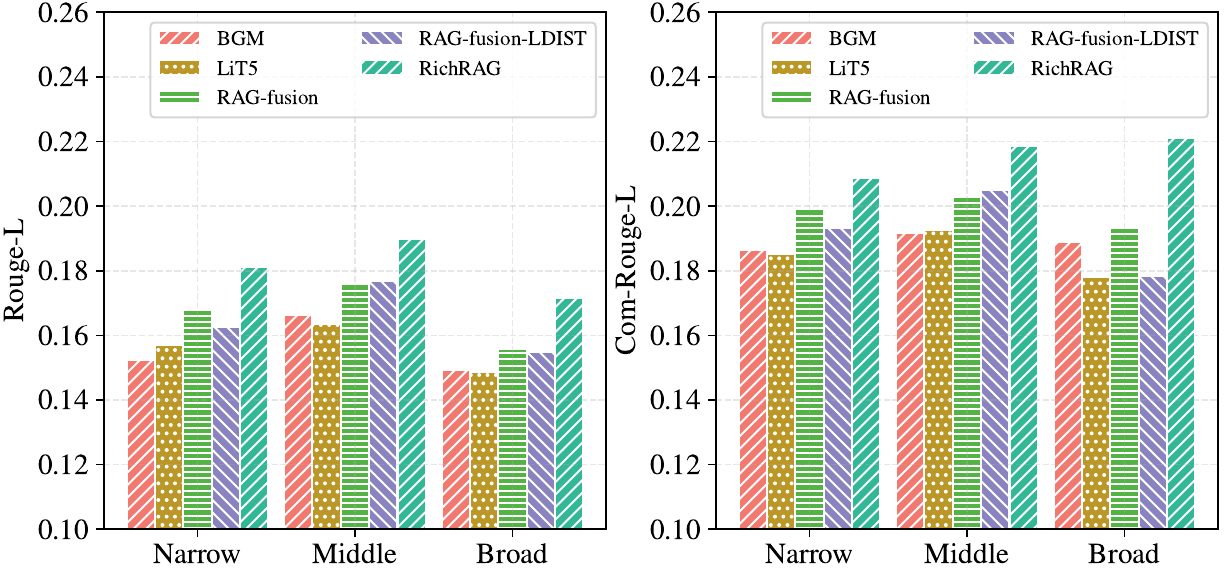}
    \caption{Subset experiments with different sub-aspect amounts.}
    \label{fig:subset}
    \vspace{-10pt}
\end{figure}

\section{Impact of Sub-aspect Amount}\label{app:subcnt}
To test the generalization of our framework with different sub-aspects numbers, which represent different search scenarios, we further split the test dataset into different sub-sets according to the number of sub-aspects. Questions with a sub-aspect amount less than two are divided into the narrow set, questions with a sub-aspect amount less than four are divided into the middle set, and the remaining questions are divided into the broad set. The models' performances on these subsets are shown in Figure~\ref{fig:subset}. Evidently, our framework outperforms all baselines on all subsets. This result verifies the robustness of our framework with diverse search scenarios. Furthermore, we find that the overall results on the broad set are worse than the remaining two sets. This phenomenon implies that the scenarios involving various potential user sub-intents are harder to handle than specific user intents, which need to be further investigated in the future.

\section{Analysis of LLMs' Preferences}\label{app:repetition}
\begin{figure}[t]
    \centering
    \includegraphics[width=0.95\linewidth]{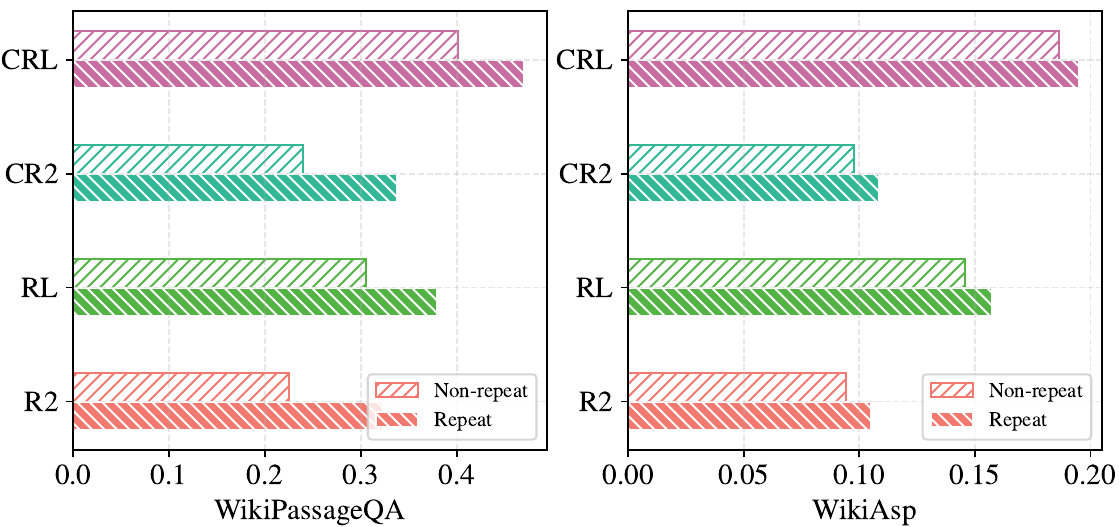}
    \caption{Comparison with and without repeated constraints}
    \label{fig:repeat}
\end{figure}

We mentioned that in the system of RAG, the downstream users of IR systems are no longer humans, but LLMs. To align the LLMs' reading preferences on the provided information, we introduce the RL stage to further capture LLMs' preferences on the order of ranking lists. Furthermore, another angle of the differences between human and LLM users is that traditional IR systems usually provide distinct documents for users and assume that users will carefully read the document containing information if she is interested in a certain document. However, such a reading habit may not be suitable for LLMs. It may be important for LLMs to repeat some important information when providing retrieved knowledge~\cite{bgm}. Such a paradigm is hard to implement by traditional ranking models based on individual relevance score sorting. However, it is easy to accomplish for our generative ranking model. Therefore, we waive the constraint of ensuring that the next ranked document has not been previously selected. The corresponding results are illustrated in Figure~\ref{fig:repeat}. Interestingly, releasing of repetition constraint could bring significant improvement to our model. The potential reason may be that repetition of important information could avoid the introduction of irrelevant information and attract more attention of LLMs to repeated important information, which enhances the LLMs' confidence in it. This enables LLMs to provide more reliable responses according to this important knowledge. 
Similar results can also be found in~\cite{wang2024domainrag}. It further confirms the importance of repetition with potential emphasis and denoising effect.

\section{Impact of Number of External Documents}\label{app:topk}
To further investigate the impact of different numbers of external knowledge on RAG performance, we vary the value of K and conduct corresponding experiments on our model and a baseline model that directly treats the top-k retrieved documents as provided references for the generator. We set the maximum value to 14 due to the limited input length of the generator. The performance trendlines of the two models are shown in Figure~\ref{fig:topk}. We further provide the trendlines of our model's improvement at each point with red lines. By comparison, we find that our model generally outperforms the baseline with different top-k numbers, which confirms the superiority of \ours{}. In addition, it is clear that the performance of the baseline usually improves with the initial increase in top-k values. This phenomenon suggests that the baseline cannot rank the valuable documents at the forefront. Therefore, important documents can only be incorporated into the generator's input when the top value becomes larger. However, our model is capable of accurately ranking the valuable documents at the top of the ranking. Consequently, even with a small k-value, it still demonstrates excellent performance. The improvement trendlines also imply that with limited external references, our model could show better response performance due to better ranking abilities. 
\section{Case Study}\label{app:case}
To validate that our proposed framework has the ability to provide rich and comprehensive responses for multi-faceted questions, we further demonstrate a case study to compare the generated responses of \ours{} and an RAG baseline that directly generates responses based on retrieved documents. In Box~\ref{box:case}, we provide the input question, its labeled sub-aspects, and predictions from different compared models. The parts of generated responses are highlighted by different colors to indicate the comprehensiveness of different results.
According to the visualized results, it is obvious that \ours{} could predict the query's sub-aspects accurately and provide rich responses that satisfy these potential intents. However, the baseline model only offers a general and incomplete response without in-depth consideration of the various underlying intentions of the user, which may implicitly impact user satisfaction. This result further proves the ability of \ours{} to identify and explicitly model the query's sub-aspects, thereby providing satisfying ranking results to enhance the richness and comprehensiveness of generated responses. This result also reveals that though LLMs have the potential ability to generate comprehensive answers, comprehensive ranking lists that correctly cover the query's sub-aspects can better stimulate LLMs' ability to generate comprehensive responses. This can also be verified by our quantitative experiments where our model exhibits the best performance among baselines.

\onecolumn

\begin{tcolorbox}[title={Box E: A case to show the richness and comprehensiveness of our model's generated responses.},label=box:case] 
\textbf{\textit{Query}}: Generate a summary about Instant Mom. \newline
\textbf{\textit{Sub-aspects}}: {\color{myorange}production}, {\color{mygreena}reception}
\tcblower
\textbf{\textit{Baseline Response}:}\newline Instant Mom is a talk show that focuses on various aspects of motherhood, including child-rearing, homemaking, and personal growth. The show features a variety of segments, including MomStop, which provides pressure-free bonding moments between mothers and their children, and Momstyle, which showcases different techniques in child rearing and domestic engineering. The show also includes 24/7, which celebrates the non-stop nature of motherhood, and Mom and Me, an inspiring tribute to the kindness and love of mothers. Additionally, the show features practical tips for homemakers and parents, as well as interviews with celebrity mothers and experts in child development. The show is certified by moms and features independent reviews and expert advice to provide a comprehensive resource for mothers.
\newline \newline
\textbf{\textit{\ours{} Response}}:\newline 
Based on the given sources, here is a summary of Instant Mom:

Instant Mom is a television show that aired on Nickelodeon from 2013 to 2015. {\color{myorange}The show was created by Howard Nemetz and starred Tia Mowry-Hardrict as Stephanie, a young woman who becomes the legal guardian of her three nieces and nephews after her sister and brother-in-law die in a car accident. The show follows the challenges and adventures of Stephanie and her new family as they navigate life and try to move forward after their loss.}

{\color{mygreena}The show received positive reviews, with Emily Ashby of Common Sense Media giving it 4 out of 5 stars and Brian Lowry of Variety comparing it to Trophy Wife. The show also featured a production report, which is a filmmaking term for a form filled out each day of production to summarize what occurred. The show was produced by Nickelodeon and was filmed in various locations in Los Angeles.}

Overall, Instant Mom is a heartwarming and humorous show that explores the complexities of family and the challenges of raising children. With its talented cast and engaging storylines, it is a great choice for families and young audiences looking for a fun and entertaining show.
\end{tcolorbox}

\begin{tcolorbox}[title={Prompt E: Annotate question aspects and split long-form answers into corresponding sub-answers},label=gpt] 
Your task is to adjust the results of query-facets mining. The query-facets are extensions of the original query in various generic perspectives, rather than some specific facts. Given a query that requires information from multiple query-facets, you should return all query-facets of the query to fully answer it query. Note that each query has at least two query-facets.   I will give you the long-form answer to the original query to help you explore query-facets based on the perspectives of its answer. But refrain from using the additional information from the answer to generate the query-facets.  Then you should segment the original long-form answer into several sub-answers that each are paired with a query-facet.  Please return each query-facet of the original query and its corresponding sub-answers. The query-facets and sub-answers should be one-to-one and returned in JSON format. You need to follow the following rules:\newline
1.  The answers are only used to help you determine the generic direction.  You mustn't generate query-facets based on the contents of answers and the facets mustn't contain the answers' additional information beyond the input query.\newline
2.  Sub-answers are constructed by segmenting the original answer, you cannot generate or reorder the original answer to create sub-answers.\newline
3.  The sub-answers should be complete. You must ensure that when the sub-answers are joined together in order, the complete original answer should be formed.\newline
4.  The generated query-facets should be sufficiently generic and contain no specific information about the sub-answers.\newline
5.  **You should ensure that generated query-facets cover all perspectives original answer.**\newline
6.  **You should ensure that all sub-answers cover all contents of the original answer.**\newline
7.  **The number of query surfaces must range from 2 to 7.**\newline
8.  **You should ensure that each query-facet is sufficiently generic and can be easily derived from the original query.**\newline
9.  **You should ensure each query-facet contains no information from the answer.** \newline
10.  **You should rewrite or combine the query-facets to be more generic if some query-facets do not meet the above requirements.**\newline
11.  The returned results should be in JSON format and contain the following key: results, which is a list of JSON data. Each item of results should contain the following keys: query-facet, and sub-answer.\newline
12.  I will give you some demonstrations, you should learn the pattern of them to mine query-facets and split sub-answers.\newline

**Demonstration**\newline
\{\textit{demonstrations}\}\newline
Query: \{\textit{query}\}\newline
Answer: \{\textit{answer}\}\newline
Results:\newline
\end{tcolorbox}


\twocolumn


\end{document}